\documentclass{article}
\usepackage[main, final]{neurips_2026}

\usepackage[utf8]{inputenc} 
\usepackage[T1]{fontenc}    
\usepackage{hyperref}       
\usepackage{url}            
\usepackage{booktabs}       
\usepackage{amsfonts}       
\usepackage{subfig}         
\usepackage{nicefrac}       
\usepackage{microtype}      
\usepackage{xcolor}         
\usepackage{amsmath}
\usepackage{amssymb}
\usepackage{mathtools}
\usepackage{amsthm}
\usepackage{hyperref}
\usepackage{url}
\usepackage{wrapfig}
\usepackage{amssymb}
\usepackage{enumitem}
\usepackage[ruled,vlined]{algorithm2e}
\usepackage{titletoc}
\usepackage[normalem]{ulem}
\usepackage{fontawesome5}   
\definecolor{linkpink}{RGB}{230,0,126}

\usepackage{amsmath,amsfonts,bm}

\def\eqref#1{equation~\ref{#1}}

\def\1{\bm{1}}

\DeclareMathAlphabet{\mathsfit}{\encodingdefault}{\sfdefault}{m}{sl}
\SetMathAlphabet{\mathsfit}{bold}{\encodingdefault}{\sfdefault}{bx}{n}

\newcommand{\bW}{{\mathbf W}}
\newcommand{\bY}{{\mathbf Y}}

\newcommand{\bH}{{\mathbf H}}
\newcommand{\bI}{{\mathbf I}}
\newcommand{\bZ}{{\mathbf Z}}
\newcommand{\bC}{{\mathbf C}}

\newcommand{\bh}{{\mathbf h}}

\newcommand{\by}{{\mathbf y}}

\newcommand{\bx}{{\mathbf x}}
\newcommand{\bb}{{\mathbf b}}

\newcommand{\bv}{{\mathbf v}}
\newcommand{\bSigma}{{\mathbf \Sigma}}
\newcommand{\byi}{{\mathbf y}_i}

\newcommand{\idx}{\ensuremath{\mathrm{ID}_{\mathrm{X}}}}
\newcommand{\idh}{\ensuremath{\mathrm{ID}_{\mathrm{H}}}}
\newcommand{\idy}{\ensuremath{\mathrm{ID}_{\mathrm{Y}}}}
\newcommand{\idp}{\ensuremath{\mathrm{ID}_{\mathrm{P}}}}

\theoremstyle{plain}
\newtheorem{theorem}{Theorem}[section]

\newtheorem{lemma}[theorem]{Lemma}

\theoremstyle{definition}

\theoremstyle{remark}
\newtheorem{remark}[theorem]{Remark}

\title{Geometric Analysis of Neural Regression Collapse via Intrinsic Dimension}

\author{
Zixuan Dong$^{1,3\dagger}$\thanks{Equal contribution. $^\dagger$Corresponding author: \texttt{zixuandong@nyu.edu}} \quad Bimarsha Adhikari$^{2*}$
  \quad George Andriopoulos$^{2*}$
  \quad Keith Ross$^{2}$\AND
  \text{\normalfont $^1$ New York University \quad
  $^2$ New York University  Abu Dhabi}\\
$^3$ SFSC of AI and DL, NYU Shanghai \\[0.6em]
  \href{https://github.com/Konokiii/neural_regression_collapse}{\textcolor{linkpink}{\faGithub\ \textbf{Code}}}}

\begin{document}

\maketitle

\begin{abstract}
  Neural multivariate regression underpins a wide range of domains such as control, robotics, and finance, yet the geometry of its learned representations remains poorly characterized. While neural collapse has been shown to benefit generalization in classification, we find that analogous collapse in regression consistently degrades performance. To explain this contrast, we analyze regression models through the lens of intrinsic dimension. Across control tasks and synthetic datasets, we estimate the intrinsic dimension of last-layer features (\idh) and compare it with that of the regression targets (\idy). Collapsed models exhibit $\idh < \idy$, leading to over-compression and poor generalization, whereas non-collapsed models typically maintain $\idh \ge \idy$. For the non-collapsed models, performance with respect to $\idh$ depends on the data quantity and noise levels. From these observations, we identify two regimes—over-compressed and under-compressed—that characterize when expanding or reducing feature dimensionality is associated with better performance. Our results provide new geometric insights into neural regression collapse and offer a practical proxy for model selection.
\end{abstract}

\section{Introduction}
\begin{wrapfigure}{r}{0.4\textwidth}
    \vspace{-12pt}
    \centering
    \includegraphics[width=\linewidth]{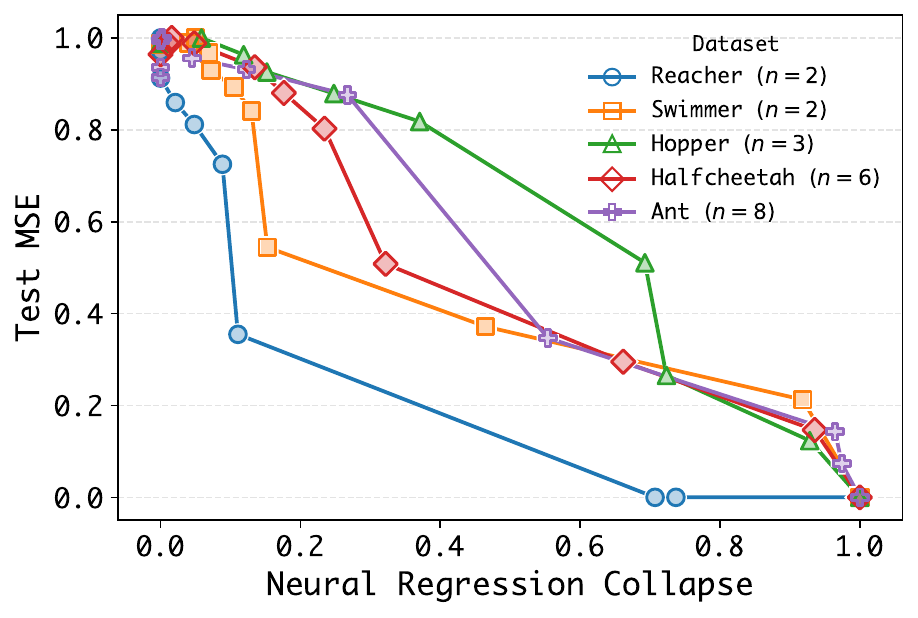}
    \caption{Neural Regression Collapse correlates with high Test MSE. The smaller the NRC value, the closer features lie to the $n$-dimensional subspace.}
    \vspace{-12pt}
    \label{fig:collapse}
\end{wrapfigure}

Neural multivariate regression has emerged as a cornerstone of modern machine learning, powering a wide spectrum of applications where the outputs are continuous and vector-valued. In imitation learning for autonomous driving, regression models predict control actions such as speed and steering angle from human driving demonstrations \citep{ross2011dagger}. In robotics, they enable agents to replicate expert trajectories \citep{chen2021dt}. In finance, regression underlies predictive analytics ranging from risk estimation to stock price forecasting \citep{gu2020asset}. 
Finally, reinforcement learning employs regression to approximate value functions, with targets derived from Monte Carlo or bootstrapped returns \citep{gallici2025pqn}. The ubiquity of regression across these domains underscores the importance of a principled understanding of the representational geometry learned by neural networks in multivariate regression tasks.  

In this work, we empirically investigate the \emph{geometric structure of neural multivariate regression}, with an emphasis on the geometry of last-layer feature vectors. Prior efforts have largely framed this problem through the lens of \emph{neural collapse}. In classification, Neural Collapse (NC) describes the emergence of a highly symmetric configuration: last-layer features converge to the vertices of a Simplex Equiangular Tight Frame (ETF), aligned with the classifier weights \citep{papyan2020prevalence}. In regression, by contrast, Neural Regression Collapse (NRC) manifests as the concentration of last-layer features within a linear subspace spanned by the top $n$ principal components of the last-layer feature matrix, where $n$ is the number of target variates. Since $n$ is typically much smaller than the feature dimension, regression collapse implies a major reduction in representational degrees of freedom \citep{andriopoulos2024prevalence}.

In this paper, we first make a key empirical observation: \emph{In contrast with classification, collapsed regression models consistently exhibit degraded generalization as compared to their non-collapsed counterparts}. Figure~\ref{fig:collapse} illustrates this phenomenon, showing high test MSE for models with highly collapsed features (low NRC metric) for five robotic locomotion tasks. 
Existing theoretical and empirical treatments of regression collapse, including the work of \citet{andriopoulos2024prevalence}, do not account for this degradation. This raises a central open question: \textbf{Why does neural collapse hinder generalization in multivariate regression, in contrast to its beneficial role in classification?}

\begin{wrapfigure}{l}{0.35\textwidth}
    \centering
    \vspace{-10pt}
    \includegraphics[width=\linewidth]{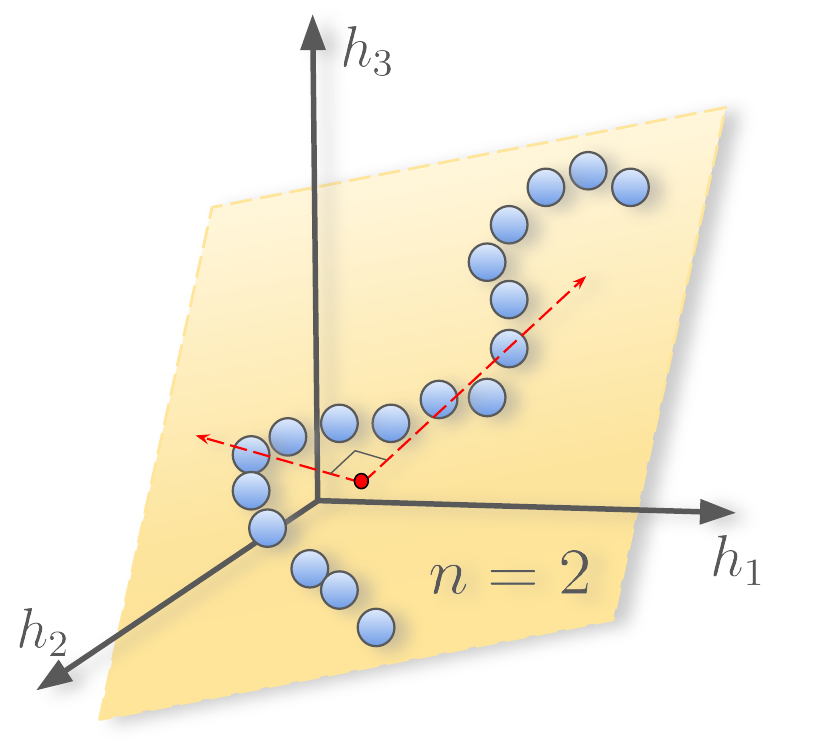}
    \caption{When the target dimension is $n=2$, the collapsed features (blue points) with intrinsic dimension of 1 can lie close to a subspace (yellow plane) spanned by the first 2 principal components (red arrows) of the features.}
    \vspace{-16pt}
    \label{fig:nrc_geometry}
\end{wrapfigure}

We address this question using \emph{intrinsic dimension (ID)}, a more refined tool than neural regression collapse for analyzing regression feature geometry. ID measures the effective dimensionality of the manifold in which data lies, capturing nonlinearities that the PCA approach of NRC cannot (Figure~\ref{fig:nrc_geometry}). We conduct a systematic study of collapsed and non-collapsed models across diverse regression tasks, including simulated robotic control and synthetic image regression. While ID has been studied in the context of neural classification \citep{ansuini2019intrinsic}, to the best of our knowledge, we are the first to systematically analyze neural regression collapse (NRC) from this perspective. 

Our findings reveal that the intrinsic dimension of the regression targets ($\idy$) acts as a critical threshold. Collapsed models typically satisfy $\idh < \idy$, where $\idh$ denotes the intrinsic dimension of the last-layer features, meaning the learned representations are over-compressed and lack the degrees of freedom needed to reconstruct the target manifold. Non-collapsed models, in contrast, typically maintain $\idh \ge \idy$. This threshold separates two regimes: an \emph{over-compressed} regime ($\idh < \idy$) where models with higher feature dimension generalize better, and an \emph{under-compressed} regime ($\idh \ge \idy$) where, in low-data or noisy settings, lower feature dimension is associated with better generalization instead. Together, these results explain why collapse is detrimental in regression and provide a practical signal for model selection. In summary, our contributions are:
\begin{itemize}[leftmargin=16pt]
    \item We provide, to the best of our knowledge, the first systematic investigation of neural regression collapse through the lens of intrinsic dimension. We show that intrinsic dimension reveals geometric structure that NRC's linear PCA methodology cannot capture, including nonlinear manifold collapse within the $n$-dimensional subspace (Section~\ref{sec:nrc_geometry}).

    \item We empirically demonstrate that the relationship between $\idh$ and $\idy$ identifies over-compressed and under-compressed regimes, explaining both why regression collapse degrades generalization and under what conditions higher or lower feature dimensionality is associated with better performance (Section~\ref{sec:generalization}). We complement these findings with a mathematical analysis that provides intuition for how regularization drives over-compression (Section~\ref{sec:mathematical_analysis}).
    
    \item Our refined geometric understanding of regression representations yields a practical diagnostic for model selection in applied regression tasks. We demonstrate that the ID gap $|\idh - \idy|$ can serve as a practical surrogate for policy evaluation in visual robotic control, motivating principled model selection that reduces costly environment rollouts (Section~\ref{sec:practical}).
\end{itemize}

\section{Related Work} \label{sec:related_work}

\textbf{NC in classification.} Neural collapse (NC) was first observed empirically by \citet{papyan2020prevalence} and has since been analyzed theoretically through the Unconstrained Feature Model \citep{mixon2020neural} and the layer-peeled model \citep{fang2021exploring}, with subsequent extensions to deep, nonlinear unconstrained-feature models \citep{tirer2022extended, sukenik2023deep}. NC has been shown to arise under diverse conditions \citep{han2021neural, yaras2022neural, zhou2022optimization, zhou2022all, zhu2021geometric} and loss functions such as label smoothing \citep{guo2024cross}. See also \citep{hong2023neural, thrampoulidis2022imbalance, yang2022inducing}. 

\textbf{NC beyond single-label classification.} \citet{li2023neural} extended NC to multi-label classification, showing that embeddings lie in the span of their label means. \citet{andriopoulos2024prevalence} formalized Neural Regression Collapse (NRC) for multivariate regression. \citet{ma2025neural} demonstrated NC in deep ordinal regression under the UFM framework, \citet{wu2024linguistic} identified ``linguistic collapse'' in large language models, and \citet{sukenik2025neural} proved NC is globally optimal in modern regularized architectures including ResNets and transformers.

\textbf{Intrinsic dimension in deep networks.} Classical local-neighborhood estimators of intrinsic dimension \citep{levina2004maximum, amsaleg2015estimating, facco2017estimating,allegra2020data} have been applied in neural settings to distinguish adversarial from natural inputs via local intrinsic dimension \citep{ma2018characterizing} and to detect untruthful LLM outputs \citep{yin2024characterizing}. Complementary topological approaches---such as Neural Persistence \citep{rieck2018neural} and graph-based invariants \citep{corneanu2019does, corneanu2020computing}---provide an alternative lens on network structure, with recent theoretical grounding via persistent homology \citep{birdal2021intrinsic}, which has also been used to regularize regression representations \citep{zhang2024topology}. Intrinsic dimension has also been studied as a complexity measure through parameter-space degrees of freedom \citep{gao2016degrees, janson2015effective}, pruning-based compressibility \citep{blier2018description}, and direct estimation \citep{ansuini2019intrinsic, li2018measuring, ma2018dimensionality, pope2021intrinsic}, with compression-based generalization bounds linking lower-dimensional representations to better generalization \citep{arora2018stronger, barsbey2021heavy, hsu2021generalization, suzuki2018spectral, suzuki2019compression, zhang2024topology}. Earlier foundational work on network redundancy includes \citet{denil2013predicting} and \citet{lecun1989optimal}.

\section{Background and Key Metrics} \label{sec:preliminary}
We consider the \emph{neural multivariate regression} problem \citep{andriopoulos2024prevalence}, with $M$ training examples $\{(\bx_i, \by_i), i=1,...,M\}$, where each input $\bx_i$ belongs to $\mathbb{R}^D$ and each target vector $\by_i$ belongs to $\mathbb{R}^n$. Considering a deep regression network of the form:
\[
f_{\theta, \bW, \bb}(\bx)=\bW \bh_{\theta}(\bx)+\bb,
\]
where $\bh_{\theta}(\cdot):\mathbb{R}^D\to \mathbb{R}^d$ is the non-linear multi-layer feature extractor, $\bW\in\mathbb{R}^{n\times d}$ represents the final linear layer in the model, and $\bb\in \mathbb{R}^n$ is the bias vector. The parameters $\theta, \bW, \bb$ are all trainable. We typically train the DNN using gradient descent to minimize the regularized $L_2$ loss:
\[
\min_{\theta, \bW, \bb} \frac{1}{2M} \sum_{i=1}^M ||f_{\theta, \bW, \bb}(\bx_i)-\by_i||_2^2+\frac{\lambda_{\text{WD}}}{2}(||\theta||_2^2+||\bW||_F^2),
\]
where $||\cdot||_2$ and $||\cdot||_F$ denote the $L_2$-norm and the Frobenius norm, respectively; and $\lambda_{\text{WD}}$ denotes the full-model weight decay coefficient.

To characterize the geometric properties of last-layer representations of neural networks in regression tasks, we consider two central metrics: the NRC1 metric \citep{andriopoulos2024prevalence} and the 2-Nearest Neighbor (2-NN) intrinsic dimension estimator \citep{facco2017estimating}.

\paragraph{Neural Regression Collapse: NRC1 metric.} \label{nrc1} In regression, neural collapse is defined by the extent to which the last-layer feature vectors collapse to a subspace spanned by their top $n$ principal components (PCs). Let $\bh_i:=\bh_{\theta}(\bx_i)$ be the feature vector associated with example $\bx_i$, $i=1,\ldots,M$.
Further let $\widetilde{\bh}_i$ be the centered normalized feature vector, that is, $\widetilde{\bh}_i:=(\bh_i-\bar{\bh})\cdot ||\bh_i-\bar{\bh}||^{-1}$ where $\bar{\bh}:=M^{-1}\sum_{i=1}^M \bh_i$. For any matrix $\bC \in \mathbb{R}^{p\times q}$ and any vector $\bv \in \mathbb{R}^p$, let $\text{proj}(\bv|\bC)$ denote the projection of $\bv$ onto the subspace spanned by the columns of $\bC$. Let $\bH_{\text{PCA}}$ be the $d\times n$ matrix with the columns consisting of the first $n$ PCs of the feature matrix $\bH = [\bh_1,\dots, \bh_M] \in \mathbb{R}^{d\times M}$. The NRC1 metric is defined as
\begin{displaymath}
\text{NRC1}:=\frac{1}{M} \sum_{i=1}^M ||\widetilde{\bh}_i-\text{proj}(\widetilde{\bh}_i|\bH_{\text{PCA}})||_2^2 .
\end{displaymath} 

\begin{wrapfigure}{r}{0.35\textwidth}
    \centering
    \vspace{-6pt}
    \includegraphics[width=\linewidth]{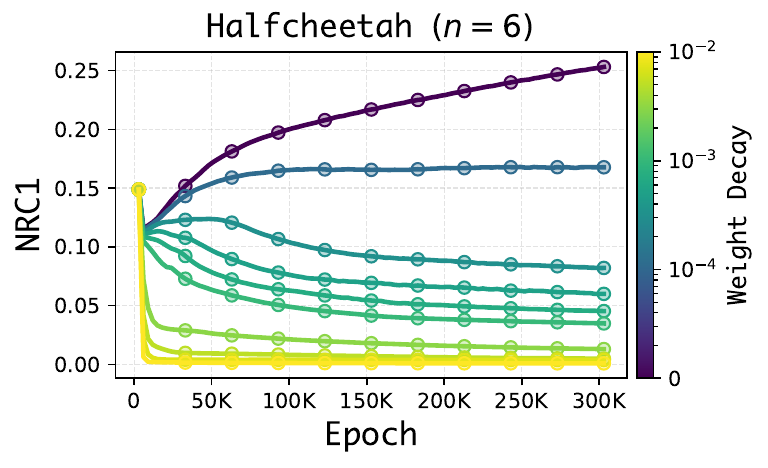}
    \caption{NRC1 decreases with stronger weight decay, leading to model collapse.}
    \label{fig:epoch-nrc}
    \vspace{-10pt}
\end{wrapfigure}

 A model is considered collapsed if NRC1 is small, indicating that the features lie almost entirely within an $n$-dimensional subspace. Non-collapsed models exhibit higher NRC1 values, differing from those of collapsed models by orders of magnitude. \citet{andriopoulos2024prevalence} demonstrated that slightly increased weight decay $\lambda_{\text{WD}}$ quickly leads to model collapse during training. Figure \ref{fig:epoch-nrc} investigates NRC1 during training with varying weight decay parameters $\lambda_{\text{WD}}$. We see that when $\lambda_{\text{WD}}$ is zero or small, there is no neural regression collapse; but if we increase the weight decay, the NRC1 geometric structure emerges. Although \citet{andriopoulos2024prevalence} attributes collapse primarily to weight decay, feature collapse is not exclusive to this regularization. In Appendix~\ref{appx:nrc1_regularization}, we empirically show that dropout regularization \citep{srivastava14adropout} and the low-rank simplicity bias induced by increasing model depth \citep{huh2023lora_bias} can also drive feature collapse, even under zero or mild weight decay.

\paragraph{Intrinsic Dimension via 2-NN Estimation.}
To uncover the finer geometry of the learned features, beyond what linear methods like PCA reveal, we turn to intrinsic dimension: the minimal number of degrees of freedom needed to describe the data without significant information loss. Unlike PCA, intrinsic dimension captures the dimensionality of nonlinear manifolds (Figure~\ref{fig:nrc_geometry}), making it a more informative geometric descriptor.

We estimate intrinsic dimension using the 2-NN estimator introduced by \citet{facco2017estimating}. For a given point, let $r_1$ and $r_2$ denote the distances to its first and second nearest neighbors; define the ratio $\mu := r_2/r_1$. Under the assumption of locally uniform data density, the cumulative distribution of the ratio follows a Pareto distribution with parameter $d$: $F(\mu) = 1 - \mu^{-d}$ for $\mu \geq 1$. The intrinsic dimension $d$ is then estimated by linear regression. 

The 2-NN estimator has been widely adopted to analyze data and representation manifolds across deep learning, including adversarial robustness~\citep{ma2018characterizing}, image classification~\citep{ansuini2019intrinsic, pope2021intrinsic}, representation correlation~\citep{basile2025id_correlation}, and continuous control via reinforcement learning~\citep{tiwari2025rl_geometry}. Beyond its ability to reliably capture nonlinear manifold structure, it is hyperparameter-free, computationally efficient, and requires only that the data density is approximately uniform within the range of each point's second neighbor, a weaker assumption than the uniformity over larger neighborhoods required by other classical estimators~\citep{levina2004maximum}. In Appendix~\ref{2nnalg}, we provide further details and intuitions, and show that our conclusions are robust to two alternative estimators, MLE~\citep{levina2004maximum} and PHD~\citep{birdal2021intrinsic}.

\section{Datasets} \label{sec:datasets}
We perform experiments on proprioceptive and visual robot locomotion datasets as well as synthetic image regression datasets, described in this section. Full experimental details are provided in Appendix~\ref{appx:exp_details}. Additionally, Appendix~\ref{appx:more_tasks} examines three more challenging datasets with varying data sizes and higher intrinsic dimensions.

\begin{table}[htbp]

\caption{Overview of datasets employed in the main body.}
\begin{center}

\scalebox{0.65}{
\begin{tabular}{ccccccc}
\toprule
\textbf{Dataset} & \textbf{Data Size} & \textbf{Input Type} &\textbf{Input Dim ($\mathrm{D}$)} & \textbf{Input ID ($\idx$)} & \textbf{Target Dim} ($n$) & \textbf{Target ID ($\idy$)}\\
\midrule
Swimmer & 20,000 & Raw state & 8 & 3.98 & 2 & 2.00\\

Reacher & 20,000 & Raw state & 11 & 3.65 & 2 & 1.99\\

Hopper & 20,000 & Raw state & 11 & 4.35 & 3  & 2.92\\

Halfcheetah & 20,000 & Raw state & 17 & 6.58 & 6  & 5.27\\

Ant & 20,000 & Raw state & 111 & 7.83 & 8 & 7.43\\

\midrule
MNIST & 50,000 & Grayscale image & 28 $\times$ 28 & 12.76 & 25 & 8.02\\
CIFAR-10 & 50,000 & RGB image & 32$\times$ 32$\times$ 3 & 27.20 & 10 & 9.51\\

\midrule
Cheetah-run & 80,000 & Stacked RGB image & $84\times84\times9$ & 8.96 & 6 & 6.00\\

\bottomrule
\end{tabular}
}

\end{center}
\label{table:dataset}
\end{table}

\paragraph{Proprioceptive MuJoCo Locomotion.} 
MuJoCo \citep{brockman2016openai, towers_gymnasium_2023} is a widely used physics simulator for continuous-control reinforcement learning. Following \citet{andriopoulos2024prevalence}, we use Reacher, Swimmer, and Hopper datasets, and additionally include the higher-dimensional Halfcheetah and Ant datasets from the D4RL benchmark \citep{fu2020d4rl}. Each dataset contains expert demonstrations with proprioceptive state inputs ($\bx_i$) and corresponding action targets ($\by_i$). States encode joint positions, angles, velocities, and angular velocities, while actions represent joint torques. We subsample the expert data to form low- and high-data regimes of 1,000 and 20,000 samples, respectively. Trained policy architectures are 3-layer MLPs with varying width, operating directly on the proprioceptive state inputs.

\paragraph{Synthetic Image Regression.}
We construct two vision-based regression tasks that differ in the amount of task-irrelevant noise present in the targets. In both cases, regression targets are generated by applying a fixed random linear projection to extracted image features.
For \emph{MNIST regression}, features are extracted from the penultimate layer of a CNN trained to over 99\% accuracy on MNIST, and then projected to 25 dimensions. Because the feature extractor is trained on the same domain, image-feature mismatches are largely suppressed, yielding clean, self-consistent targets.
For \emph{CIFAR-10 regression}, features are extracted using a ResNet-18 pretrained on ImageNet and \emph{not} fine-tuned on CIFAR-10, and then projected to 10 dimensions. This domain mismatch causes the projected targets to retain instance-specific, task-irrelevant variation, resulting in noisy targets.
For both tasks, regression models are 3-layer MLPs with varying width, operating on the same image tensors used to generate synthetic targets.

\paragraph{Visual Continuous Control.}
Visual control is particularly challenging because each frame provides only partial information about the system state, effectively forming a Partially Observed Markov Decision Process (POMDP)~\citep{yarats2022drq_v2}. The model must infer underlying transition dynamics and identify salient visual features directly from high-dimensional pixel inputs. We use the Cheetah-run dataset~\citep{lu2023vd4rl}, a visual control benchmark derived from the DeepMind Control Suite~\citep{tassa2018deepmindcontrolsuite}, which also builds on the MuJoCo engine but differs from the proprioceptive tasks above in that inputs are raw RGB image observations. Following standard practice in visual control, we stack three consecutive frames to form $84 \times 84 \times 9$ input tensors, providing temporal context for inferring velocities and dynamics. Targets correspond to continuous control commands ($n=6$), and we employ up to 80K expert demonstrations from the dataset. During training, the model architecture consists of a 4-layer CNN image encoder followed by a 3-layer MLP policy head.

\section{Intrinsic Dimension Refines NRC Geometry} \label{sec:nrc_geometry}
NRC1 measures the degree to which last-layer features collapse to an $n$-dimensional linear subspace, and as shown in Figure~\ref{fig:epoch-nrc}, a small amount of weight decay often suffices for such collapse to occur. However, NRC1 does not reveal whether the features further collapse into lower-dimensional nonlinear manifolds, a limitation shared by other spectrum-based metrics such as effective rank (Appendix~\ref{appx:spectral_metrics}). To explore this question, we measure the intrinsic dimension of the last-layer features, denoted $\idh$, via the 2-NN estimator.

\begin{figure}[h]
    \centering
    \includegraphics[width=\textwidth]{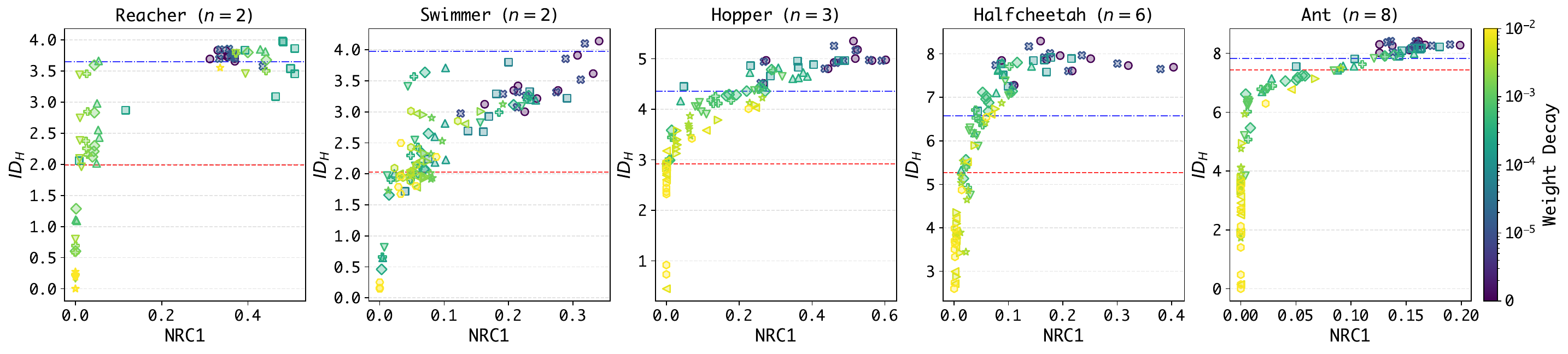}
    \caption{Relationship between NRC1 and intrinsic dimension of the last-layer features $\idh$. Dots correspond to models trained with different architectures and weight decay parameters, with colors denoting the degree of weight decay. The horizontal red dashed line is drawn at $\idy$, the intrinsic dimension of the targets. The blue line is drawn at $\idx$, the intrinsic dimension of the inputs.}
    \label{fig:nrc1_id}
\end{figure}

Figure~\ref{fig:nrc1_id} presents scatter plots of $\idh$ versus NRC1 for five MuJoCo datasets. The critical value $\idy$, the intrinsic dimension of the targets, denoted by the dashed-red lines in Figure \ref{fig:nrc1_id}, is consistently upper-bounded by $n$, the dimension of the targets. For highly collapsed models with near-zero NRC1, the intrinsic dimension of last-layer features satisfies $\idh \le \idy$ and continues to shrink even as NRC1 saturates at zero. That is, \emph{collapsed models learn features that lie on increasingly lower-dimensional nonlinear manifolds within an $n$-dimensional linear subspace}. So, although the NRC1 metric is useful in understanding the linear-subspace structure of the last-layer features in collapsed models, it is inadequate at uncovering this more refined geometric structure. 

\begin{figure}[h]
    \centering
    \begin{minipage}[c]{0.3\textwidth}
        \centering
        \includegraphics[width=\linewidth]{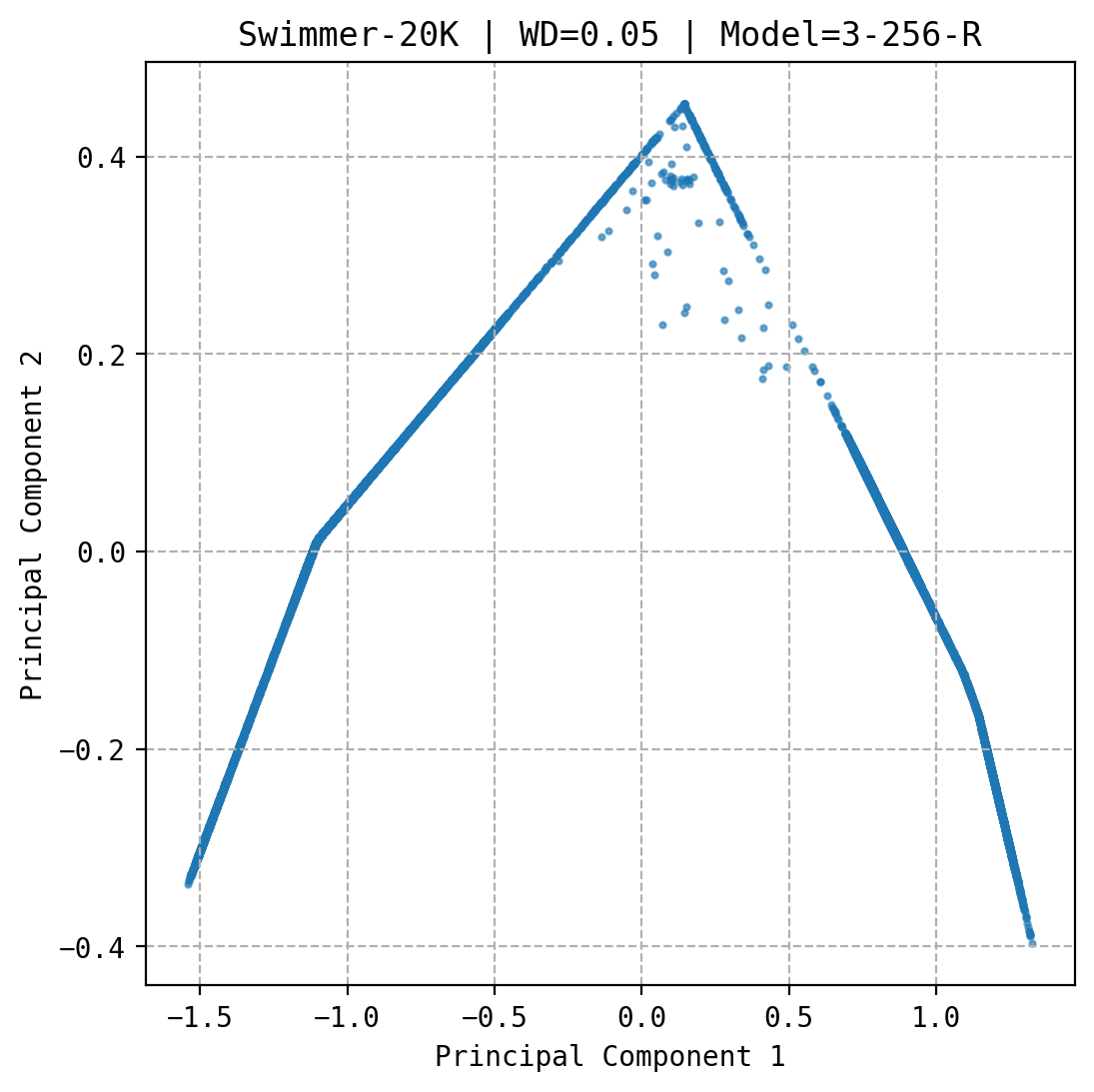}
    \end{minipage}
    \hfill
    \begin{minipage}[c]{0.318\textwidth}
        \centering
        \includegraphics[width=\linewidth]{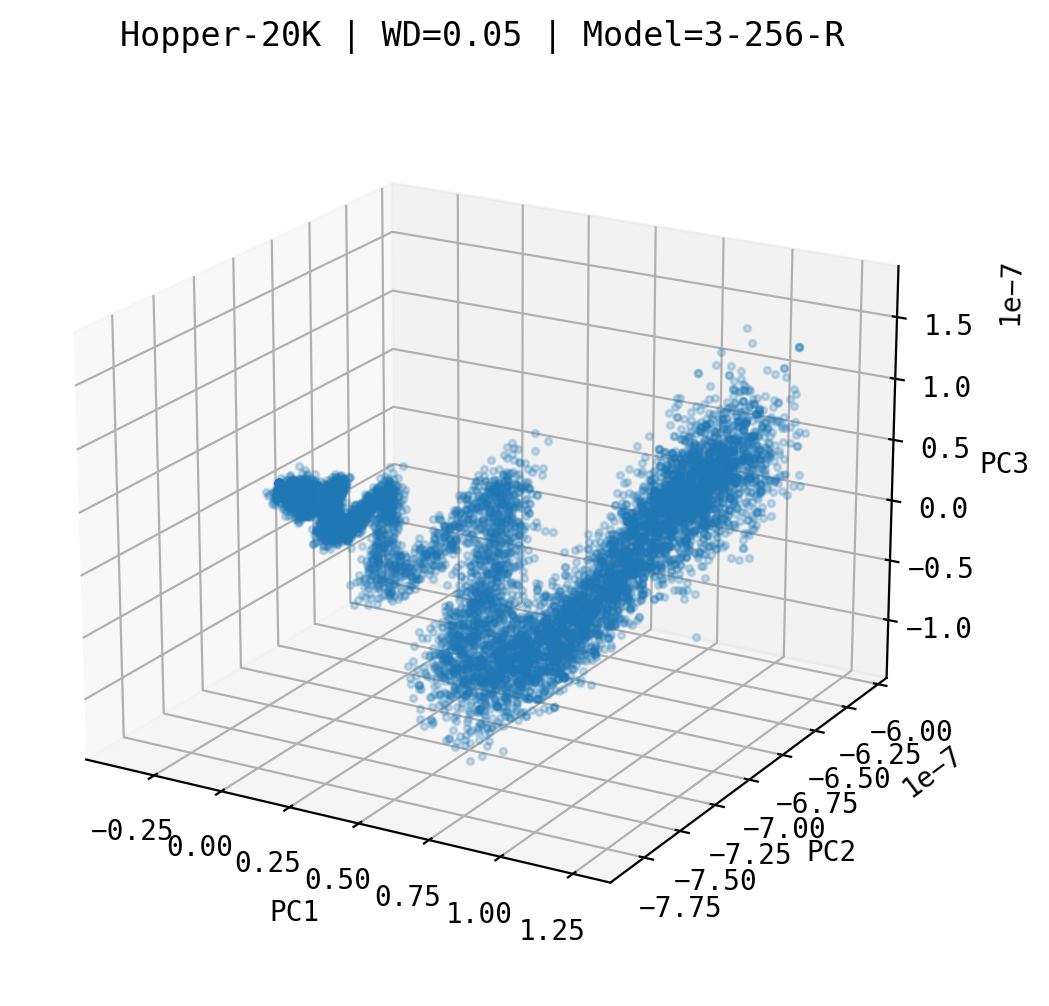}
    \end{minipage}
    \hfill
    \begin{minipage}[c]{0.31\textwidth}
        \centering
        \small
        \setlength{\tabcolsep}{4.5pt}
        \begin{tabular}{ccc}
            \toprule
            $k$ & Global PCA & Local PCA \\
            \midrule
            6 & $3.4\!\times\!10^{-3}$ & $2.8\!\times\!10^{-5}$ \\
            5 & $7.5\!\times\!10^{-3}$ & $7.9\!\times\!10^{-5}$ \\
            4 & $8.4\!\times\!10^{-2}$ & $3.5\!\times\!10^{-4}$ \\
            3 & $4.9\!\times\!10^{-1}$ & $1.5\!\times\!10^{-3}$ \\
            \bottomrule
        \end{tabular}
    \end{minipage}
    \caption{Nonlinear structure of collapsed feature manifolds. \textbf{Left, Center:} PCA projections of collapsed 1D feature manifolds for Swimmer ($n=2$) and Hopper ($n=3$), showing visibly curved structures. \textbf{Right:} Global vs.\ local PCA projection error for collapsed Halfcheetah features ($\idh = 3.6$, $n=6$). Local errors (projection onto the top-$k$ PCs of 64-nearest-neighbor feature matrix) are orders of magnitude smaller, confirming locally flat but globally curved geometry.}
    \label{fig:nonlinear_manifold}
\end{figure}

To further verify this nonlinear structure, Figure~\ref{fig:nonlinear_manifold} visualizes PCA projections of collapsed 1D features in Swimmer ($n=2$) and Hopper ($n=3$), revealing visibly curved manifolds that cannot be spanned by a single linear subspace. For the higher-dimensional Halfcheetah dataset, we examine a collapsed feature manifold with $\idh = 3.6 < n = 6$. We compare global PCA projection errors (onto the top-$k$ PCs of the full feature matrix) with local PCA projection errors (onto the top-$k$ PCs of each point's 64 nearest neighbors). As shown in the table of Figure~\ref{fig:nonlinear_manifold}, the global error increases sharply from $k=5$ to $k=4$, while the local error jumps from $k=4$ to $k=3$. This indicates that the nearly-4D feature manifold is embedded in a 5D linear subspace and cannot itself be linear. These results echo the analysis of curved feature manifolds in classification by \citet{ansuini2019intrinsic}.

In contrast, non-collapsed models with higher NRC1 values satisfy $\idh > \idy$, and their intrinsic dimension increases monotonically with NRC1, making the two metrics qualitatively interchangeable in this regime. In other words, \emph{non-collapsed models learn features whose manifold dimension exceeds that of the targets}. Intrinsic dimension thus offers two concrete advantages over NRC1: it reveals a soft threshold at $\idy$ separating two geometrically distinct regimes, and it quantifies the degree of collapse ($\idh$) across the full range of NRC1 values. For the remainder of this paper, we therefore focus on intrinsic dimension. Appendix~\ref{appx: id_evolution} further analyzes the evolution of intrinsic dimension during training.

\section{Intrinsic Dimension Predicts NRC Generalization} \label{sec:generalization}

Having now investigated the relationship between NRC and intrinsic dimension, we now examine what insights intrinsic dimension can provide about NRC generalization. Particularly, we will explore why generalization error increases as $\idh$ (and hence as NRC1) decreases, as seen in Figure~\ref{fig:collapse}. As we discuss in more detail at the end of this section, this property is in contrast with classification, for which performance typically improves when neural collapse becomes stronger.

\begin{figure}[htpb]
    \centering
    \includegraphics[width=\textwidth]{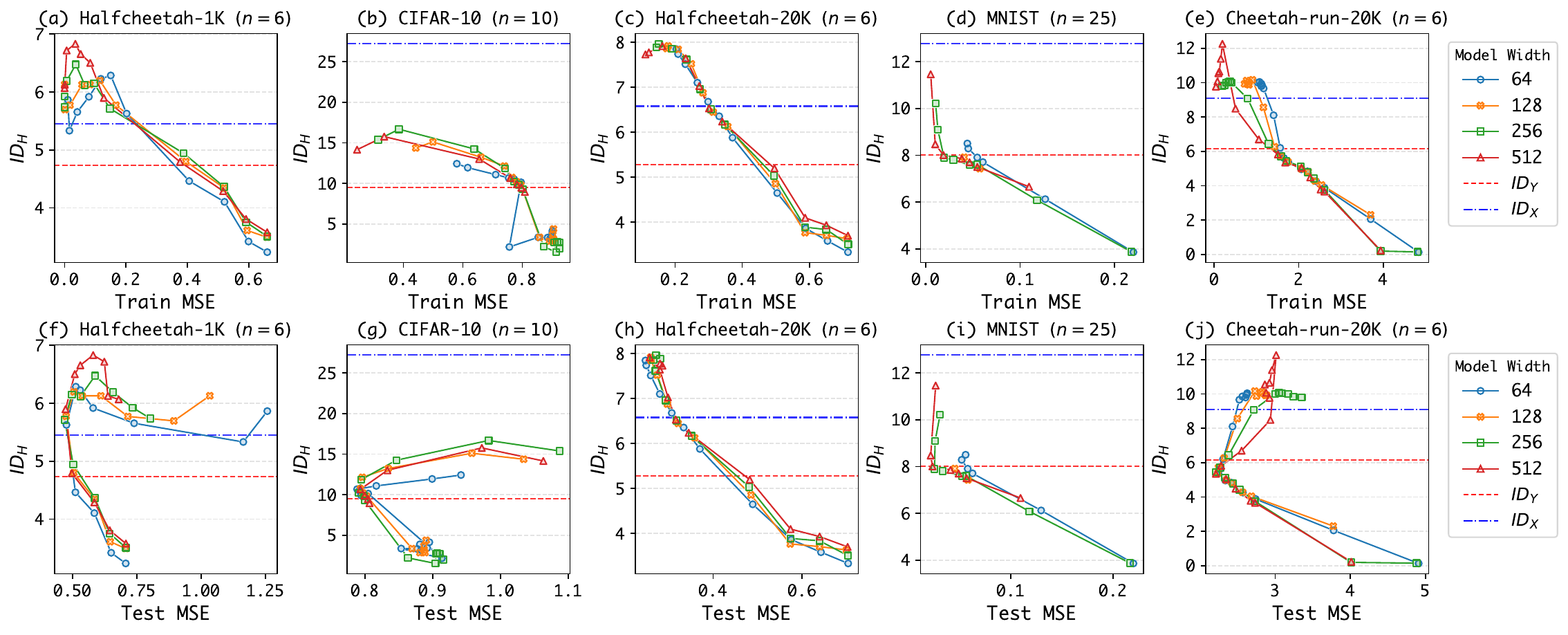}
    \caption{Generalization and Intrinsic Dimension for Halfcheetah, CIFAR-10, MNIST, and Cheetah-run datasets.}
    \label{fig:generalization}
    \vspace{-10pt}
\end{figure}

Figure \ref{fig:generalization} shows the relationship between $\idh$ and both training and test MSE for five datasets: Halfcheetah-1K, Halfcheetah-20K, CIFAR-10, MNIST, and Cheetah-run-20K datasets. The corresponding figures for other MuJoCo datasets are in Appendix~\ref{appx:generalization}.

\paragraph{Train MSE decreases when \(\mathbf{\idh}\) increases.} This trend is evident in the first row of Figure~\ref{fig:generalization}. To explain this, from Figures~\ref{fig:epoch-nrc} and ~\ref{fig:nrc1_id} we have known stronger regularization reduces $\idh$. Theorems 4.1 and 4.3 in \citet{andriopoulos2024prevalence} also tell that stronger regularization reduces the dimension of the linear subspace containing the feature manifold. Thus, by reducing $\idh$, the trained features tend to get squashed onto a lower-dimensional and more curved manifold, similar to the "crowding problem" described by \citet{maaten2008tsne}. A global linear layer $\bW$, performing only an affine transformation, cannot "unbend" such a manifold. Thus, as $\idh$ decreases, it becomes more difficult for $\bW \bH +\bb$ to accurately match $\bY$ (which lies on its own curved manifold), explaining why train MSE decreases when $\idh$ increases.

\paragraph{Test MSE with respect to \(\mathbf{\idh}\) behaves differently according to its relationship to \(\mathbf{\idy}\).} The second row of Figure \ref{fig:generalization} shows fundamental differences between collapsed and non-collapsed models:

\begin{itemize}[leftmargin=16pt]
    \item $(\idh < \idy)$:
    In this regime, the collapsed model's features are confined to a manifold whose intrinsic dimension is lower than that of the targets. This \textit{over-compression} intuitively means the last-layer features lack information essential for reconstructing and generalizing beyond the full target manifold, and Section \ref{sec:mathematical_analysis} will present a mathematical formulation of this claim. In this regime, models with higher last-layer intrinsic dimension, obtained for example with weaker regularization or a different architecture, generalize better. We can now answer the question posed in the Introduction: Why does neural collapse hinder generalization in multivariate regression? The explanation simply follows from $(1)$ the monotonic relationship between NRC1 and $\idh$; and $(2)$ the reconstruction inability as just described. 

    \item $(\idh \geq \idy)$: We distinguish these non-collapsed models between two cases:

    \begin{itemize}[leftmargin=12pt]
        \item {\em (i) Low-data or noisy-target tasks.}
            In this case (Figs.~\ref{fig:generalization} (f), (g), (j)), the test MSE plots exhibit surprising U-shaped dependence on $\idh$, with a minimum error near $\idh = \idy$. Why does test MSE increase with $\idh$ when $\idh > \idy$? To explain this, we note that when the amount of training data is small, or when the targets are noisy, $f_\theta$ learns a feature manifold whose intrinsic dimension is higher than that of the true feature manifold due to the stronger negative effect of outliers. Then, the extra dimensions in the feature manifold are used to predict training sample-specific noise, leading to overfitting the training set \citep{ma2018dimensionality, ansuini2019intrinsic}. This overfitting is exacerbated when regularization is reduced, or equivalently, when $\idh$ is increased, leading to higher test MSE.
    
        \item {\em (ii) High-data and low-noise tasks.}
            In this case, test MSE follows the same trend as train MSE, decreasing monotonically with $\idh$ (Figs.~\ref{fig:generalization} (h),(i)). To explain, we note that with a large amount of training data and low target noise, $f_\theta$ can fit the training data closely while maintaining smoothness to avoid overfitting, and consequently, the training feature manifold $\bH_{\text{train}}$ is likely to resemble the test feature manifold $\bH_{\text{test}}$. 
    \end{itemize}
    
\end{itemize}

\begin{table}[ht]
\caption{ Key Takeaways for Generalization.}
\centering
\scalebox{1.0}{
\begin{tabular}{cccccccc}
\toprule
\textbf{Regime} & \textbf{ID} & \textbf{Typical behavior} \\
\midrule
Over-compressed & $\idh < \idy$ & Underfitting with large train and test MSE \\
Balanced & $\idh \approx \idy$ &
Sweet spot in low-data and noisy tasks \\
Under-compressed & $\idh > \idy$ &
Benign overfitting with enough clean data\\
\bottomrule
\end{tabular}
}
\label{table:gen}
\end{table}

Table~\ref{table:gen} provides the key takeaways concerning generalization and intrinsic dimension. In Appendix~\ref{appx:comparison_with_classification}, we review the generalization of neural collapse for classification tasks and contrast it to our study for multivariate regression tasks.

\subsection{Practical Guideline: Intrinsic Dimension as a Surrogate for Policy Evaluation} \label{sec:practical}

\begin{wrapfigure}{r}{0.45\textwidth}
\vspace{-16pt}
    \centering
    \includegraphics[width=\linewidth]{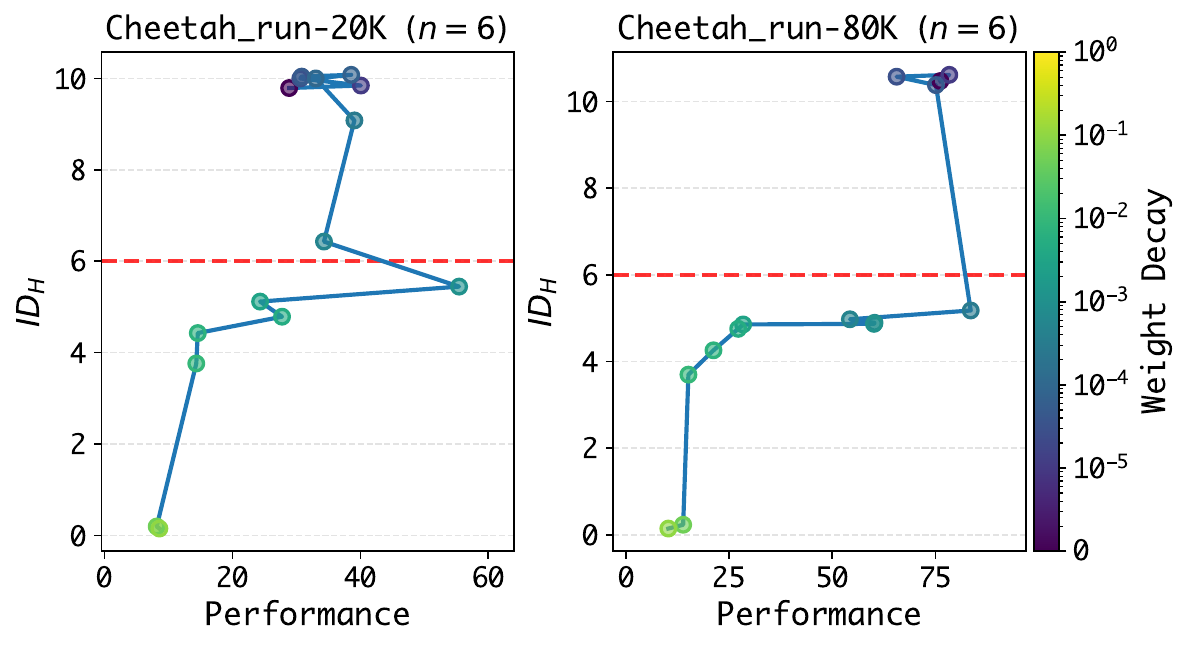}
    \caption{Relationship between $\idh$ and normalized model performance ($\in[0,100]$; the higher, the better). The horizontal red dashed line represents $\idy$. }
    \label{fig:rl_evaluation}
    \vspace{-10pt}
\end{wrapfigure}

Our findings regarding NRC geometry have direct implications for evaluating real-world control models, where deployment-time evaluation is expensive, unsafe, and resource-limited~\citep{levine2020offline_rl}. We demonstrate this on the Cheetah-run visual control task introduced in Section~\ref{sec:datasets}, where we train policies via behavior cloning on 20K and 80K expert demonstrations and evaluate them by executing the learned policy in the DeepMind Control Suite simulator~\citep{tassa2018deepmindcontrolsuite}.

Figure~\ref{fig:rl_evaluation} plots the relationship between $\idh$ and \emph{actual} normalized policy performance (ranging from 0 to 100). Consistent with the test MSE analysis, performance exhibits an inverted U-shape, peaking near $\idh = \idy$, which mirrors the ID--generalization relationship identified in Section~\ref{sec:generalization}. Crucially, validation MSE alone is an unreliable proxy for policy quality: small MSE differences can correspond to large variations in control performance (cf.\ Fig.~\ref{fig:generalization} (j) and Fig.~\ref{fig:rl_evaluation}). In contrast, the gap between $\idh$ and $\idy$ provides a complementary signal that consistently predicts both low test MSE and high policy performance.

This observation motivates an actionable model selection strategy. Our analysis shows that well-generalizing models either maximize $\idh$ in the high-data regime or achieve $\idh \approx \idy$ in the low-data and noisy-target regime. Since the regime is rarely known in advance, we retain both types of models: those with the largest $\idh$ and those with the smallest ID gap, $|\idh - \idy|$. Together, they form a compact candidate pool that contains the potentially best-performing policies regardless of the regime, serving as a principled selection criterion. Compared to exhaustive hyperparameter grid search, this approach offers two advantages. First, it avoids missing optimal configurations in regions where the learned feature geometry changes abruptly across regimes. For instance, in the first row of Figure~\ref{fig_appx:id_evolution} (Reacher), increasing weight decay by as little as $2 \times 10^{-4}$ can shift a model into the over-compressed regime. Second, it reduces dependence on costly real-environment rollouts for policy evaluation, which in real robotic systems are inherently risky and resource-constrained.

While these results constitute preliminary evidence on a simulated benchmark, they point toward a promising direction: automated, efficient hyperparameter selection via ID gap minimization, complementing or partially replacing expensive deployment-time evaluation.

\section{Mathematical Intuition of Regression Collapse}
\label{sec:mathematical_analysis}

To provide a principled explanation for the empirical phenomena observed in Section~\ref{sec:generalization}, we present a mathematical characterization of regression collapse. We first describe how weight decay induces a dimensional reduction in the feature space (Section~\ref{sec:why_collapse}) and subsequently formalize why this necessitates a reconstruction error (Section~\ref{sec:why_fail}).

\subsection{Why Weight Decay Causes Collapse}
\label{sec:why_collapse}

We analyze the effect of regularization through the Unconstrained Feature Model (UFM) framework adopted in prior work~\citep{mixon2022ufm,andriopoulos2024prevalence}, which treats the last-layer feature matrix as a free optimization variable rather than the output of a parameterized feature extractor. In this section, we slightly abuse some notations introduced in Section~\ref{sec:preliminary}. Let the feature matrix be $\bH \in \mathbb{R}^{d \times M}$. Given a centered target matrix $\tilde{\bY} \in \mathbb{R}^{n \times M}$ with empirical covariance $\bSigma = \frac{1}{M}\tilde{\bY}\tilde{\bY}^\top \in \mathbb{R}^{n\times n}$ (assuming $\bSigma$ is full rank with eigenvalues $\lambda_1 \geq \cdots \geq \lambda_n > 0$), the UFM minimizes:
\begin{equation}
\min_{\bW, \bH} \;\frac{1}{2M} \|\bW\bH - \tilde{\bY}\|_F^2 \;+\; \frac{\lambda_W}{2}\|\bW\|_F^2 \;+\; \frac{\lambda_H}{2}\|\bH\|_F^2,
\label{eq:ufm}
\end{equation}
where $\bW \in \mathbb{R}^{n \times d}$ is the final linear layer; and $\lambda_W$, $\lambda_H$ are separate $L_2$ regularization parameters.\footnote{Practical regression objective defined in Section~\ref{sec:preliminary} applies the same weight decay regularization to all model parameters.} Without regularization ($\lambda_W = \lambda_H = 0$), global minimizers satisfy $\bW\bH = \tilde{\bY}$ and decompose as $\bH = \bW^\dagger \tilde{\bY} + (\bI_d - \bW^\dagger \bW)\bZ$ for arbitrary $\bZ \in \mathbb{R}^{d \times M}$~\citep{andriopoulos2024prevalence}. The term $\bW^\dagger \tilde{\bY}$ lies in the at-most-$n$-dimensional row space of $\bW$, while $(\bI_d - \bW^\dagger \bW)\bZ$ spans the $(d{-}n)$-dimensional null space of $\bW$, allowing $\bH$ to occupy up to $d$ dimensions. With regularization, our following result shows that this null-space freedom is eliminated.

\begin{theorem}[Limiting Solution Structure; informal, see Appendix~\ref{appx:ufm_collapse}]
\label{thm:limiting_informal} Let $(\bW_\lambda, \bH_\lambda)$ be the global minimizer of~\eqref{eq:ufm} and suppose $\lim_{\lambda_H \to 0, \lambda_W \to 0}(\lambda_H/\lambda_W) = k > 0$. Then, as regularization gradually vanishes, $\bW_\lambda \bH_\lambda \to \tilde{\bY}$ and the limiting features $\bH_0 = \lim \bH_\lambda$ satisfy $\bH_0 = \bW_0^\dagger \tilde{\bY}$, the minimum-norm solution with $\bZ = 0$. In particular, $\bH_0$ lies entirely within the $n$-dimensional row space of $\bW_0$.
\end{theorem}

Thus, even infinitesimally small weight decay biases features toward the minimum-norm solution, collapsing $\bH$ from the $d$-dimensional ambient space onto an $n$-dimensional subspace. Since typically $d \gg n$, this represents a massive dimensional reduction.

\paragraph{From Neural Regression Collapse to ID Over-compression.}
Theorem~\ref{thm:limiting_informal} explains why features concentrate in an $n$-dimensional linear subspace, but does not yet explain why $\idh$ can fall \emph{below} $\idy$. This transition is governed by the product $c := \lambda_W \lambda_H$. When $c < \lambda_n$ (the smallest eigenvalue of $\bSigma$), the regularized solution maps $\tilde{\bY}$ to $\bH_\lambda$ via an invertible affine transformation, preserving intrinsic dimension, e.g., $\idh = \idy$. However, when $c \geq \lambda_n$, the solution reduces $\text{rank}(\bH_\lambda) < n$ and breaks the invertibility. As a consequence, affine congruency is not guaranteed to hold, allowing $\idh$ to drop below $\idy$, entering the over-compressed regime. The formal characterization is provided in Remark~\ref{rem:ufm_affine_congruency} (Appendix~\ref{appx:ufm_collapse}).

This analysis highlights the diagnostic value of intrinsic dimension in practice. The UFM parameter $c = \lambda_W \lambda_H$ is not directly observable, as it arises from a simplified theoretical abstraction rather than an explicit practical training objective. However, intrinsic dimension serves as an empirical proxy: the transition from $\idh \approx \idy$ to $\idh < \idy$ signals that the effective regularization strength has crossed the critical threshold $c = \lambda_n$, placing the model in the over-compressed regime where affine congruency between features and targets breaks down.

\subsection{Why Collapsed Models Fail to Generalize}
\label{sec:why_fail}

The over-compression identified above creates an unavoidable reconstruction error, which we formalize by treating the final layer as a smooth map $g$ from the feature manifold to the target space.

\begin{theorem}[Non-surjectivity of Over-compressed Maps]
\label{thm:non_surjectivity}
Let $\mathcal{M}_H$ be a smooth $m$-dimensional feature manifold and $\mathcal{N}_Y$ be a smooth $n$-dimensional target manifold, with $m < n$. A smooth map $g: \mathcal{M}_H \to \mathcal{N}_Y$ cannot be surjective; specifically, the image $g(\mathcal{M}_H)$ has Lebesgue measure zero in $\mathcal{N}_Y$.
\end{theorem}

As proven in Appendix~\ref{appx:non_surjectivity} via Sard's Theorem, the condition $m < n$ in the over-compressed regime implies that the set of all possible model predictions is a proper subset of the target manifold. Geometrically, there will always be points on the target manifold that lie outside the model's predictive reach, making perfect reconstruction mathematically impossible. This explains why over-compressed models consistently exhibit high test MSE across all experiments (Figures~\ref{fig:generalization}, \ref{fig_appx:generalization_appx_1} and \ref{fig_appx:generalization_appx_2}).

\section{Limitations and Future Work}
\label{sec:limitation}
\paragraph{Limitations.} Our theoretical results explain why weight decay induces collapse and why over-compressed models cannot fully reconstruct the target manifold, but a complete characterization of the relationship between intrinsic dimension and generalization is not yet available. These results are derived under the Unconstrained Feature Model (UFM), which treats the last-layer features as free optimization variables and thus abstracts away the feature extractor, network depth, and architecture. Our empirical evidence linking $\idh$ to generalization is correlational: $\idh$ varies through the interventions that naturally induce NRC (weight decay, dropout, and depth), which may also affect other aspects of training. Accordingly, the ID gap should be read as a predictive signal for model selection rather than as a quantity whose direct manipulation is guaranteed to improve generalization.
\paragraph{Extending the theory beyond the UFM.} A natural path for future work is to follow the progression in the classification literature from the UFM to deep, nonlinear unconstrained-feature models and beyond (Section~\ref{sec:related_work}). For regression, this requires replacing the cross-entropy loss with the squared loss and discrete classes with a continuous target manifold. Such an analysis could account for the layer-wise ID reduction we observe in collapsed models (Appendix~\ref{appx: id_evolution}), clarifying where along depth over-compression arises and how architecture interacts with the $\idy$ threshold. Deriving generalization bounds that incorporate NRC and ID would further ground the empirical relationships we report.

\paragraph{Characterizing data regimes that favor a small ID gap.} Our analysis concerns the regularized regression problem, in which $\idh$ varies through the interventions that naturally induce NRC (weight decay, dropout, and depth). A complementary line of work instead imposes intrinsic dimension directly as a training objective for regression, using differentiable persistent-homology estimators \citep{birdal2021intrinsic} to match the intrinsic dimension and topology of the features to those of the targets \citep{zhang2024topology}. Our results suggest that whether such matching is beneficial depends on the data regime: a small ID gap $|\idh - \idy|$ is favored in low-data or noisy-target settings, whereas performance continues to improve with larger $\idh$ in high-data, low-noise settings. Characterizing more precisely which properties of the data, such as sample size, target noise, and target-manifold geometry, favor a small ID gap is an important direction for both regularization-induced and explicitly ID-regularized regression.

\paragraph{Identifying implicit modulators of NRC and ID.} Weight decay, dropout, and depth are not designed to control intrinsic dimension, yet each of them shifts $\idh$ and, through it, generalization. Other common regularizers and algorithmic design choices, such as normalization layers, data augmentation, the choice of optimizer, learning-rate schedules, batch size, and early stopping, may play a similar implicit role. Systematically studying which of these choices modify NRC and $\idh$, in which direction, and with what effect on generalization is both interesting and practically important: it would reveal when standard training recipes unintentionally push models into the over-compressed regime, and would make the ID gap a diagnostic for tuning them. Finally, exploring whether intrinsic dimension provides similar insights for other learning paradigms, such as generative modeling, reinforcement learning, and multi-task learning, could offer a unifying geometric perspective across domains.

\section{Conclusion}
In this paper, we provided a systematic geometric analysis of neural regression collapse through the lens of intrinsic dimension. We showed that regression collapse corresponds to an over-compressed regime ($\idh < \idy$) that discards information essential for reconstructing target manifolds, explaining why collapse, beneficial in classification, is detrimental in regression. For non-collapsed models, generalization behavior depends on whether the task is low-data/noisy or high-data/low-noise. These findings yield a practical guideline: over-compression ($\idh < \idy$) signals poor generalization, and models with the largest $\idh$ or the smallest ID gap $|\idh - \idy|$ form a candidate pool of well-generalizing models regardless of the data regime, providing a principled criterion for model selection in applied multivariate regression tasks such as robotic control.

\newpage
\section*{Acknowledgments}
This work was supported in part by NYU Abu Dhabi Center for Artificial Intelligence and Robotics and the Center for Interdisciplinary Data Science and AI, funded by Tamkeen under the Research Institute Award CG010. It was also partially supported by the Shanghai Frontiers Science Center of Artificial Intelligence and Deep Learning at NYU Shanghai. Computational resources were provided in part by NYU IT High-Performance Computing resources and services.

\bibliography{neurips_2026}
\bibliographystyle{neurips_2026}

\newpage
\appendix
\addcontentsline{toc}{section}{Appendices}

\section*{\LARGE \bfseries Appendix Contents}
\label{sec:appendix_toc}

\par\vspace{0.5em}
\noindent\rule{\textwidth}{0.6pt}
\vspace{0.3em}

\startcontents[appendix]
\printcontents[appendix]{l}{1}{\setcounter{tocdepth}{2}}

\par\vspace{0.3em}
\noindent\rule{\textwidth}{0.6pt}
\vspace{0.5em}

\clearpage

\section{Experiment Details} \label{appx:exp_details}

\begin{table}[htbp]

\caption{Overview of datasets employed in the main body.}
\begin{center}

\scalebox{0.65}{
\begin{tabular}{ccccccc}
\toprule
\textbf{Dataset} & \textbf{Data Size} & \textbf{Input Type} &\textbf{Input Dim ($\mathrm{D}$)} & \textbf{Input ID ($\idx$)} & \textbf{Target Dim} ($n$) & \textbf{Target ID ($\idy$)}\\
\midrule
Swimmer & 20,000 & Raw state & 8 & 3.98 & 2 & 2.00\\

Reacher & 20,000 & Raw state & 11 & 3.65 & 2 & 1.99\\

Hopper & 20,000 & Raw state & 11 & 4.35 & 3  & 2.92\\

Halfcheetah & 20,000 & Raw state & 17 & 6.58 & 6  & 5.27\\

Ant & 20,000 & Raw state & 111 & 7.83 & 8 & 7.43\\

\midrule
MNIST & 50,000 & Grayscale image & 28 $\times$ 28 & 12.76 & 25 & 8.02\\
CIFAR-10 & 50,000 & RGB image & 32$\times$ 32$\times$ 3 & 27.20 & 10 & 9.51\\

\midrule
Cheetah-run & 80,000 & Stacked RGB image & $84\times84\times9$ & 8.96 & 6 & 6.00\\

\bottomrule
\end{tabular}
}

\end{center}
\label{table:dataset}
\end{table}

\subsection{MuJoCo experiments} \label{appx: mujoco}
MuJoCo (Multi-Joint dynamics with Contact) is a physics engine designed for research in robotics, biomechanics, and animation, providing fast and accurate simulations of systems involving complex contact dynamics. It balances physical realism with computational efficiency, enabling reliable modeling of robot–environment interactions \citep{towers2024gymnasium}. Environments involved in this work include:

\begin{itemize}
    \item \textbf{Reacher}: A two-jointed robotic arm tasked with moving its tip to a randomly generated target in a 2D plane. 
    \item \textbf{Swimmer}: A chain-like robot with three body segments connected by two rotors, aiming to propel itself forward in 2D as quickly as possible.
    \item \textbf{Hopper}: A one-legged, four-part robot that seeks to hop forward at maximum speed in 2D.
    \item \textbf{HalfCheetah}: A planar, bipedal robot with a torso and two legs, each consisting of two joints. It aims to run forward as quickly as possible along a 2D track by coordinating its leg movements.
    \item \textbf{Ant}: A quadrupedal robot with four legs and multiple joints, designed to move in a 3D plane. Its goal is to walk or run forward efficiently, despite the challenge of balancing and coordinating many degrees of freedom. Although Ant's state space has 111 dimensions, 84 of the dimensions related to external contact forces are always zeros in the dataset. Thus, the effective input dimension is 27.
\end{itemize}

\begin{figure}[htb]
    \centering
    \includegraphics[width=0.6\linewidth]{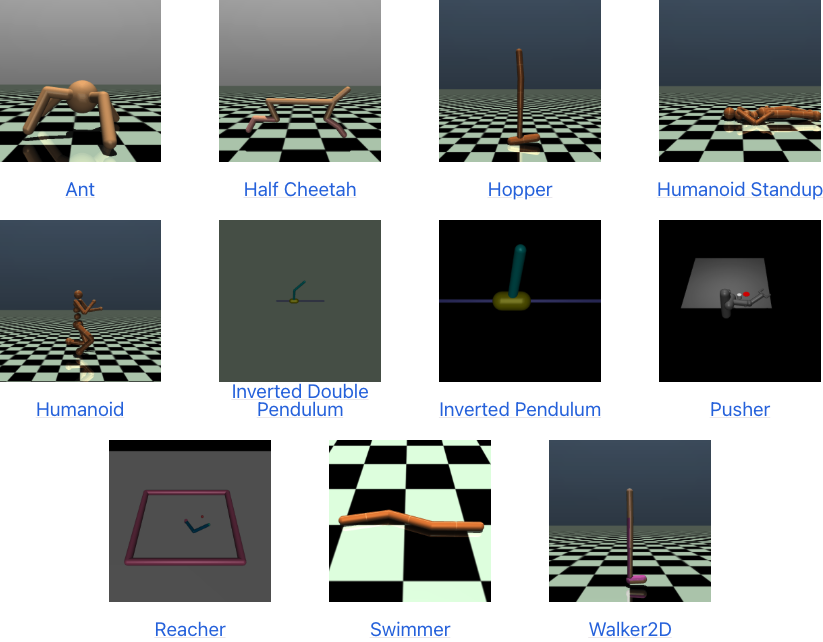}
    \caption{Screenshot of various MuJoCo environments \citep{towers2024gymnasium}.}
    \label{fig:placeholder}
\end{figure}

All environments introduce stochasticity by perturbing a fixed initial state with Gaussian noise. Their state spaces combine positions of body and joint with corresponding velocities. Control is achieved by applying joint torques, which serve as the actions. Expert datasets are generated by first training policies through online reinforcement learning \citep{fu2020d4rl,gallouedec2024jack} until high performance, then executing these policies to produce trajectories of states $\bx_i$ and actions $\byi$. Here, $\bx_i$ encodes robot positions, joint angles, velocities, and angular velocities, while $\byi$ denotes the applied joint torques.

An episode of expert demonstration has a length of 50 for Reacher, and it has a length of 1,000 for all other environments. Thus, by taking 20,000 data points from each expert dataset, the regression model learns from at least 20 complete trajectories to clone the expert's behavior. For evaluation, we retain a subset of the full validation dataset, keeping the number of data points at 20\% of the training data size. For small datasets (1K) used in Figures~\ref{fig:generalization}, \ref{fig_appx:generalization_appx_1} and \ref{fig_appx:generalization_appx_2}, the test datasets contain 1,000 unseen samples. 

There is no absolute threshold for what constitutes "low-data" versus "high-data," as this depends on problem complexity and data structure. In our experiments, we operationally define this distinction using datasets of 1,000 samples versus 20,000 samples—a 20-fold difference that produces qualitatively different generalization behavior. In the low-data regime, the training set provides sparse coverage of the true data manifold, introducing sampling artifacts such as spurious correlations and outlier effects that do not reflect the true underlying distribution. Models trained on such small datasets, particularly without sufficient regularization, tend to memorize these sample-specific patterns rather than learning generalizable structure. In the high-data regime with 20,000 samples, the training set provides denser coverage, the effect of sampling artifacts diminishes, and the empirical distribution more closely approximates the true underlying distribution.

\begin{table}[h!]
\caption{All hyperparameter settings involved for experiments on MuJoCo datasets. Each figure employs a subset of possible hyperparameter combinations.}
\label{table:mujoco_params}
\vskip 0.15in
\begin{center}
\scalebox{0.69}{%
\begin{tabular}{cll}
\toprule
    & \textbf{Hyperparameter} & \textbf{Value}  \\
\midrule
    & Number of hidden layers   & $\{1,2,3,4,5\}$ \\
  Model Architecture  & Hidden layer dimension & $\{64, 128, 256, 512, 1024\}$  \\ 
    & Activation function & ReLU  \\ 
    & Number of linear projection layer ($\bW$) & 1 \\
\midrule
    & Epochs & $3\times10^5$ (20K-datasets) \\
    & & $5\times10^6$ (1K-datasets) \\
    & Batch size & 4096 (20K-datasets) \\
    & & 1000 (1K-datasets) \\
    & Optimizer & SGD \\
    & Learning rate & $1\times10^{-2}$ \\
Training    & Weight decay & $\{0, 10^{-5}, 10^{-4}, 3\times10^{-4}, 5\times10^{-4}, 7\times10^{-4}, 10^{-3}, 3\times10^{-3} \}$, Reacher \\
    &   & $\{0, 10^{-5}, 1/3/5/7\times10^{-4}, 1/3/5/7\times10^{-3}, 10^{-2}, 3\times10^{-2} \}$, Otherwise \\
    & Seed & 0 \\
    & Compute resources & NVIDIA A100 40GB \\
    & Number of CPU compute workers & 4 \\
    & Requested compute memory & 16 GB \\
    & Average training time per model & 20 hours\\
    
\bottomrule
\end{tabular}%
}
\end{center}
\end{table} 

Table~\ref{table:mujoco_params} summarizes all model hyperparameters and experimental settings for MuJoCo datasets. A subset of possible hyperparameter combinations is used for each figure:
\begin{itemize}
    \item Figure~\ref{fig:collapse} plots the min-max normalized Test MSE as a function of the min-max normalized NRC1 values for the model architecture 3-256 (3 hidden layers and 256 hidden units) and all possible weight decay values. 
    
      \item Figure~\ref{fig:nrc1_id} establishes the relationship between NRC1 and $\idh$. Each subplot includes all weight decays listed in Table~\ref{table:mujoco_params}. And each weight decay is combined with 9 model architectures in \{3-64, 3-128, 3-256, 3-512, 3-1024, 1-256, 2-256, 4-256, 5-256\}.

      \item Figs.~\ref{fig:generalization}, \ref{fig_appx:generalization_appx_1} and \ref{fig_appx:generalization_appx_2} empirically reveal how generalization ability is affected by $\idh$. We focus on a single model depth of 3, and vary the model width among $\{64, 128, 256, 512\}$. For each model architecture, we evaluate all possible weight decay values listed in Table~\ref{table:mujoco_params}.

    \item Figure~\ref{fig_appx:id_evolution} depict how intrinsic dimension evolves for each network layer. The model architecture is fixed at 3-256 and the title of each subplot annotates the weight decay value.
    
    \item Figure~\ref{fig:6x3_idp1} and \ref{fig:6x3_idp2} follow the same experimental setup as Figs.~\ref{fig:generalization}, \ref{fig_appx:generalization_appx_1} and \ref{fig_appx:generalization_appx_2}, but emphasize on the comparison between $\idh$ and $\idp$.

      \item Figure~\ref{fig_appx:nrc1_wd} record NRC1 values along the training process. The model architecture is fixed to 3-256 for all datasets. And we show 10 weight decay values in $\{0, 0.0001, 0.0003, 0.0005, 0.0007, 0.001, 0.003, 0.005, 0.007, 0.01\}$.
\end{itemize}

\subsection{MNIST/CIFAR10 experiments} \label{appx: vision}
The regression models for both the MNIST and CIFAR-10 tasks were trained across a spectrum of hyperparameters to thoroughly investigate the effects of architecture and regularization on the learned representations. The specific settings for model architecture, optimizer, and other training parameters are detailed in Table~\ref{table:mnist_cifar_params}.

\begin{table}[h!]
  \caption{All hyperparameter settings involved for experiments on MNIST and CIFAR-10 datasets.}
  \label{table:mnist_cifar_params}
  \vskip 0.15in
  \begin{center}
  \scalebox{0.7}{
  \begin{tabular}{cll}
  \toprule
      & \textbf{Hyperparameter} & \textbf{Value}  \\
  \midrule
      & Number of hidden layers  & 3 \\
    Model Architecture  & Hidden layer dimension & $\{32, 64, 128, 256, 512\}$  \\
      & Activation function & ReLU  \\
  \midrule
      & Epochs & 200 \\
      & Batch size & 64 \\
      & Optimizer & Adam \\
      & Learning rate & $\{1 \times 10^{-3}, 5 \times 10^{-3}\}$ \\
  Training    & Weight decay (MNIST) & $\{0, 10^{-5}, 10^{-4}, 3 \times 10^{-4}, 5 \times 10^{-4}, 7 \times 10^{-4}, 10^{-3}, 3 \times
  10^{-3}, 7 \times 10^{-3}\}$ \\
      & Weight decay (CIFAR-10) & $\{0, 10^{-5}, 10^{-4}, 3 \times 10^{-4}, 5 \times 10^{-4}, 7 \times 10^{-4}, 10^{-3}\}$ \\
      & Seed & 0 \\
      & Compute resources & NVIDIA A100 80GB \\
      & Average training time per model & 2 hours\\

  \bottomrule
  \end{tabular}
  }
  \end{center}
  \end{table}

We define noisy-target tasks as tasks where the regression targets contain information that causes models to learn patterns that do not generalize to unseen data. The term "noise" here does not refer to random measurement error or label corruption, but rather to information that varies between semantically similar examples due to instance-specific characteristics rather than reflecting the true underlying function.

The synthetic MNIST regression task represents a low-noise setting. The feature extractor is a CNN trained specifically on MNIST until achieving over 99\% classification accuracy, creating a self-consistent target-generation pipeline. During training, the CNN learns to discard instance-specific variations (such as stroke thickness, slight rotations, or pixel-level noise) that are irrelevant for predicting the target label, retaining only the semantically meaningful information. As a result, the mapping from MNIST images to projected features is smooth and well-aligned with the data domain, enabling models to generalize effectively (Figs.~\ref{fig:generalization} (d),(i)).

In contrast, the synthetic CIFAR-10 regression task is a noisy-target setting due to domain mismatch. CIFAR-10 images are processed through a ResNet-18 pretrained on ImageNet to extract features, which are then projected to 10-dimensional targets. The ImageNet encoder captures fine-grained visual details—texture patterns, color distributions, edge structures—optimized for ImageNet's 1,000 classes. When applied to CIFAR-10's 32×32 images, these features encode not only semantic content but also instance-specific characteristics: two images of cars may differ in color, lighting, rendering artifacts, or background elements, all of which significantly influence the extracted features.
Models can fit these instance-specific components during training, achieving low training error. However, at test time, new images from the same semantic classes exhibit different instance-specific details. The learned mappings for these non-generalizable components do not transfer, resulting in poor generalization. This is evident in Figs.~\ref{fig:generalization} (b),(g), where models achieve low training MSE but high test MSE, characteristic of overfitting to target noise.

\subsection{Visual Control Experiments} \label{appx:vd4rl}
 
Visual D4RL~\citep{lu2023vd4rl} provides offline reinforcement learning benchmarks derived from the DeepMind Control Suite~\citep{tassa2018deepmindcontrolsuite}, where the observations consist of raw RGB pixel inputs rather than proprioceptive states. This introduces additional challenges compared to the state-based MuJoCo tasks described in Appendix~\ref{appx: mujoco}, as the model must extract task-relevant features directly from high-dimensional visual observations while discarding irrelevant visual details.
 
\paragraph{Environment.} We use the Cheetah-run environment, where a planar bipedal robot must run forward as fast as possible. Following standard practice in visual continuous control~\citep{yarats2022drq_v2}, we stack three consecutive $84 \times 84 \times 3$ RGB frames to form $84 \times 84 \times 9$ input tensors, providing temporal context for inferring velocities and dynamics. The target actions are 6-dimensional continuous control commands ($n = 6$). The environment constitutes a Partially Observed Markov Decision Process (POMDP), as individual frames do not fully capture the system state.
 
\paragraph{Datasets.} Expert demonstration datasets are generated by first training policies via online reinforcement learning until high performance, then collecting trajectories of image observations and corresponding actions. We consider two data regimes: 20,000 and 80,000 training samples. Test datasets contain 20,000 samples.
 
\paragraph{Model architecture.} The model consists of two components: a CNN image encoder and an MLP policy network. The encoder contains four convolutional layers, all with 32 output channels and $3 \times 3$ kernels (stride 2 for the first layer, stride 1 for the remaining three), each followed by ReLU activation. Input images are normalized to $[-0.5, 0.5]$, and the encoder outputs a flattened representation of dimension $32 \times 35 \times 35 = 39{,}200$. This representation is passed through a trunk consisting of a linear projection to 50 dimensions, followed by layer normalization and a Tanh activation. The policy head is a 3-hidden-layer MLP with hidden dimension 256 and ReLU activations, followed by a final linear projection layer $\mathbf{W}$ that maps to the 6-dimensional action space. During evaluation, actions are squashed through a $\tanh$ nonlinearity to enforce the $[-1, 1]$ action bounds. The last-layer features $\mathbf{h}$ used for intrinsic dimension and NRC1 analysis correspond to the output of the 3-layer MLP policy head (before the final linear projection $\mathbf{W}$).
 
\paragraph{Training.} Models are trained using the Adam optimizer with a fixed learning rate of $3\times10^{-4}$. The maximum number of training steps is set to $10 \times$ the training data size (e.g., $8 \times 10^5$ steps for the 80K dataset). Weight decay is applied to the network and is varied across a wide range of values:
$
\lambda_{\text{WD}} \in \{0,\; 10^{-5},\; 3 \times 10^{-5},\; 5 \times 10^{-5},\; 7 \times 10^{-5},\; 10^{-4},\; 3 \times 10^{-4},\; 5 \times 10^{-4},\; 7 \times 10^{-4},\; 10^{-3},\; 3 \times 10^{-3},\; 5 \times 10^{-3},\; 7 \times 10^{-3},\; 10^{-2},\; 5 \times 10^{-2},\; 10^{-1}\}
$.
 
\paragraph{Evaluation.} In addition to train and test MSE, we evaluate the learned policies (of 256 hidden dimensions) by executing each of them in the DeepMind Control Suite simulator for $N_{\text{episodes}} = 100$ episodes. Policy performance is reported as a normalized score $\in [0, 100]$ averaged over $N_{\text{episodes}}$ evaluations, obtained by dividing the raw evaluation score by 10 \citep{tarasov2023rebrac}.

\section{Comparison between Regression and Classification} \label{appx:comparison_with_classification}

Previous work on manifold learning for neural classification has demonstrated that the intrinsic dimension of the last hidden layer is negatively correlated with generalization ability. In particular, models achieving lower intrinsic dimension in the penultimate layer were found to exhibit superior test accuracy, with the lowest intrinsic dimension-model attaining the highest top-5 accuracy, see Section 3.2 and Figure 4 in \citet{ansuini2019intrinsic}. 
Additionally, \citet{papyan2020prevalence} connect neural collapse to robust decision boundaries,
\citet{galanti2021role} demonstrate that collapse patterns improve few-shot and transfer learning,
and \citet{li2022understanding} show the degree of collapse in downstream representations strongly predicts transfer accuracy. Complementing these empirical results, there are also theoretical results showing the benefits of neural collapse for classification \citep{gao2023towards,wang2023towards}.

In regression, however, our findings indicate a more nuanced picture. We demonstrated the existence of a ``soft" threshold at $\idy$, which delineates distinct generalization regimes. In the under-compressed regime with low-data tasks and high-noise tasks, lower $\idh$ is associated with better generalization, consistent with the monotonic complexity-performance paradigm observed in classification. However, in the over-compressed regime and in the under-compressed regime with high-data tasks and low-noise tasks, the opposite holds: higher $\idh$ is associated with better generalization, a phenomenon absent in classification tasks. Thus, in regression, generalization performance depends non-monotonically on the relationship between the learned feature manifold and the intrinsic dimension of the targets.

\section{Intrinsic Dimension Estimation} \label{2nnalg}
The intrinsic dimension (ID) of a dataset is the minimum number of coordinates needed to represent the data faithfully. If data points lie on or near a $d$-dimensional manifold $\mathcal{M}$ embedded in $\mathbb{R}^D$ with $d \ll D$, then $d$ is the intrinsic dimension. For example, a circle in 3D space has $d = 1$ and a sphere surface in 10D has $d = 2$.

A critical distinction exists between PCA dimensionality and intrinsic dimension. Consider a 1D spiral embedded in $\mathbb{R}^{10}$ parameterized by $t$. The spiral winds through space with substantial variance across all 10 axes, requiring multiple PCA components to capture the signal. Yet, the intrinsic dimension is exactly $d=1$, because specifying a single scalar value, such as the arc length from the origin, is sufficient to uniquely locate any point on the curve. This illustrates that curved or folded manifolds can require many linear directions to approximate while having low intrinsic dimension. Figure~\ref{fig:nrc_geometry} demonstrates this for neural regression: collapsed features lie near a 2D linear subspace (yellow plane) yet occupy a nonlinear 1D manifold within it.

Our main analysis uses the 2-NN estimator (Appendix~\ref{appx:2nn}). In Appendix~\ref{appx:estimator_comparison}, we verify that our conclusions are robust to this choice by comparing it with two widely used alternatives, MLE~\citep{levina2004maximum} and PHD~\citep{birdal2021intrinsic}.

\subsection{The 2-NN Estimator} \label{appx:2nn}

We estimate ID using the 2-NN estimator \citep{facco2017estimating}, which exploits a fundamental geometric property: in a $d$-dimensional space, the probability of finding neighbors within a given distance scales with dimension $d$. The key insight is to consider not absolute distances, but the ratio $\mu = r_2/r_1 \geq 1$ of the second to first nearest-neighbor distances. Remarkably, under the assumption of locally uniform density (density approximately constant within the range of the second neighbor), this ratio has a distribution that depends only on the intrinsic dimension $d$, with the local density completely canceling out. Specifically, the cumulative distribution function of $\mu$ is \[F(\mu) = 1 - \mu^{-d}, \quad \mu \geq 1\] 
This property makes the estimator robust to density variations, since we never need to estimate the density itself. Taking logarithms yields the linear relationship \[\log(1 - F(\mu)) = -d \log \mu\]

The 2-NN algorithm estimates $d$ by computing $\mu_i$ for each point, constructing the empirical CDF $F_{\text{emp}}$, and performing linear regression on the transformed coordinates $\{(\log \mu_i, -\log(1 - F_{\text{emp}}(\mu_i)))\}$. The requirement of local uniformity only within the second-neighbor distance is much weaker than global uniformity, making the estimator practical for real datasets with varying density and curvature. 

\begin{algorithm}[H]
\caption{2-NN Intrinsic Dimension Estimation}
\label{alg:2nn}

\KwIn{Dataset $\mathcal{X} = \{\bx_i\}_{i=1}^M$.}
\KwOut{Estimated intrinsic dimension $\hat d$.}

\For{$i \gets 1$ \KwTo $M$}{
  Compute Euclidean distances to the first and second nearest neighbors, $r_1(\bx_i)$ and $r_2(\bx_i)$\;
  
  Compute ratio $\mu_i \gets r_2(\bx_i) / r_1(\bx_i)$\;
}

Sort the ratios such that $\mu_{\sigma(1)} \le \mu_{\sigma(2)} \le \cdots \le \mu_{\sigma(M)}$\;

\For{$i \gets 1$ \KwTo $M$}{
  Assign empirical CDF value $F_{\text{emp}}(\mu_{\sigma(i)}) \gets \frac{i}{M}$\;
}

Construct the coordinate set for regression:
\[
\mathcal{S} \gets \Big\{ \big(\log \mu_{\sigma(i)},\; -\log \!\big(1 - F_{\text{emp}}(\mu_{\sigma(i)})\big)\big) \Big\}_{i=1}^{M-1}
\]

Fit a line through the origin to $\mathcal{S}$ using least squares\;

\Return Slope of the fitted line ($\hat d$)\;

\end{algorithm}

\paragraph{Robustness of the 2-NN estimator.} The robustness of the 2-NN estimator is well justified in the literature and has been widely adopted across various domains~\citep{facco2017estimating, ansuini2019intrinsic, basile2025id_correlation}. It features: (1) the capability to capture intrinsic dimension on nonlinear manifolds; (2) fast computation; and (3) hyperparameter-free estimation. Moreover, as suggested by \citet{facco2017estimating}, we follow the standard block analysis procedure to evaluate the robustness of our ID estimates. The motivation behind block analysis is to distinguish true manifold structure from noise: intrinsic dimensions corresponding to meaningful geometry should remain stable across sample scales, whereas noise-driven dimensions diminish under subsampling. Concretely, we repeatedly estimate ID on progressively smaller random subsets of the data and observe how the estimate changes; a stable plateau indicates genuine intrinsic structure rather than a sampling artifact.

\begin{figure}[htb]
    \centering
    \subfloat[Non-collapsed Halfcheetah features.]{%
        \includegraphics[width=0.48\linewidth]{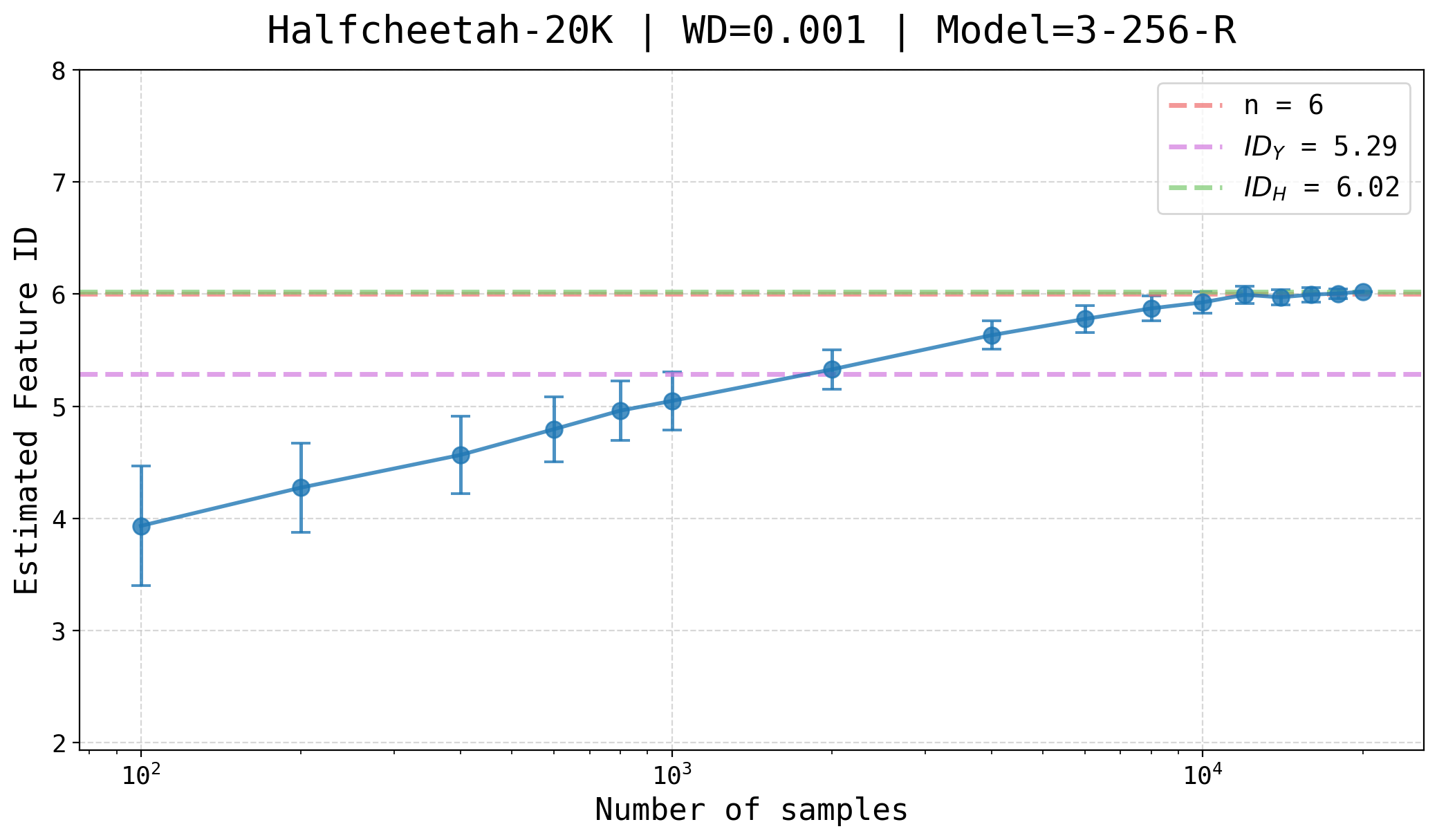}%
    }
    \hfill
    \subfloat[Collapsed Ant features.]{%
        \includegraphics[width=0.48\linewidth]{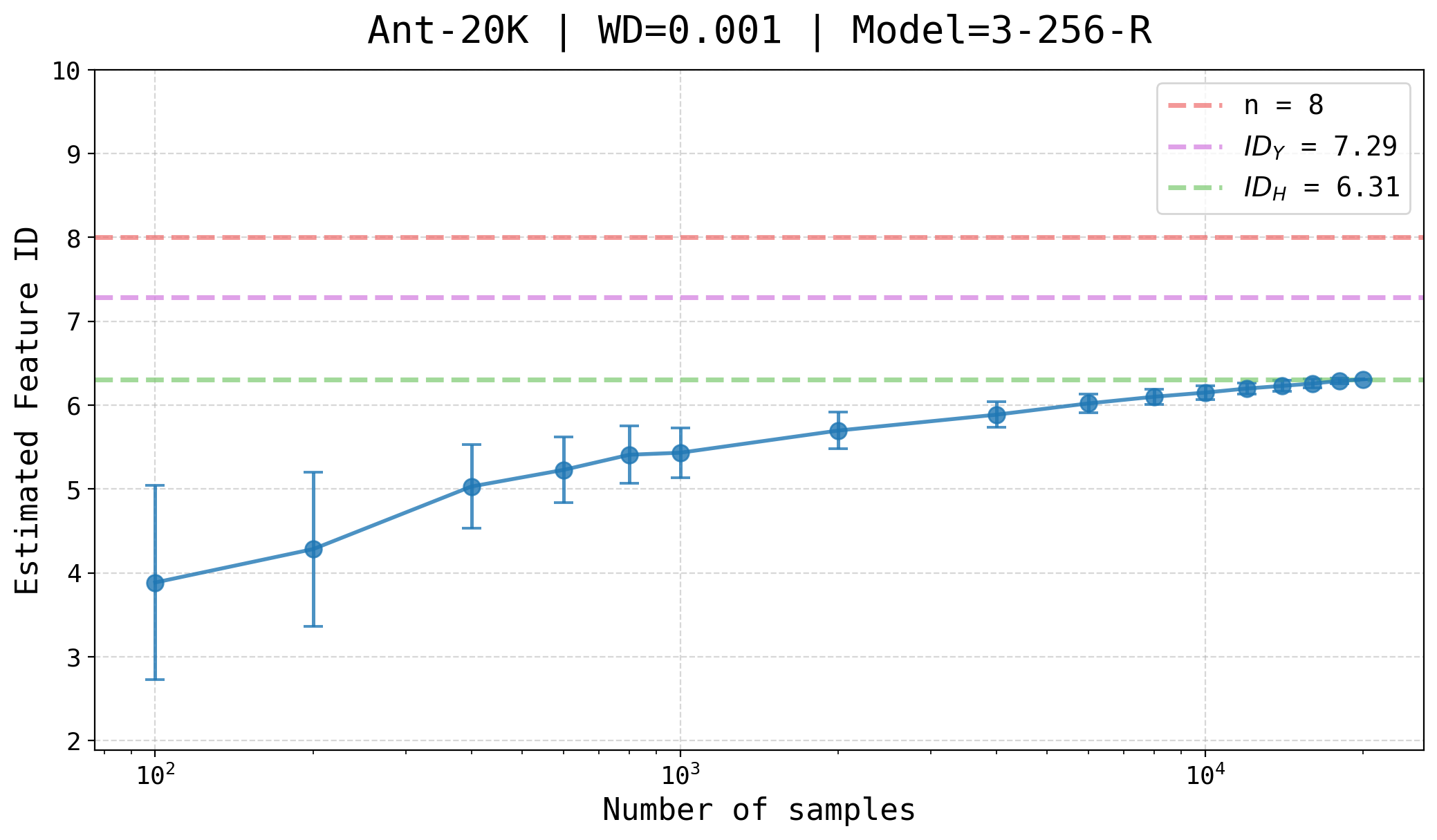}%
    }
    \caption{Block analysis of the 2-NN intrinsic dimension estimator. Each point shows the mean estimated ID over 100 random subsamples of the given block size. Both cases exhibit a stable plateau for block sizes above approximately 10{,}000, confirming the reliability of the ID estimates at our default subsample size of 18{,}000.}
    \label{fig:block_analysis}
\end{figure}

We demonstrate block analyses for two representative cases in Figure~\ref{fig:block_analysis}: (a) non-collapsed Halfcheetah last-layer features and (b) collapsed Ant last-layer features. Block sizes range from 100 to 20{,}000, and for each block size, we subsample 100 times from the training dataset and report the mean and standard deviation of the estimated ID. In both cases, we observe that the ID estimate converges to a stable plateau once the block size exceeds approximately 10{,}000 samples, with diminishing variance as the block size increases. This confirms that the reported ID values reflect true manifold structure rather than finite-sample noise, and that our default subsample size of 18{,}000 lies well within the stable regime.

\subsection{Robustness to the Choice of Estimator} \label{appx:estimator_comparison}
To verify that our conclusions do not hinge on the 2-NN estimator, we repeat the ID estimation with two widely used alternatives. The maximum likelihood estimator (MLE) of \citet{levina2004maximum} is, like 2-NN, a local estimator. Around each sample, it models the number of neighbors falling within a growing radius as a Poisson process whose rate scales with the radius according to the intrinsic dimension, assuming that the density is approximately constant within the neighborhood. Maximizing the resulting likelihood yields a closed-form local dimension estimate computed from the distances to the sample's $k$ nearest neighbors, and the global estimate averages these local estimates over all samples. Compared with 2-NN, which uses only the first two neighbors, MLE pools information from a larger neighborhood, trading some locality for lower variance at the cost of introducing the neighborhood size $k$ as a hyperparameter.

The persistent homology dimension (PHD) of \citet{birdal2021intrinsic} instead takes a global, topological view of the point cloud. It computes the total lifetime of the 0-dimensional persistent homology, which coincides with the total edge length of the minimum spanning tree over the samples. For data lying on a $d$-dimensional manifold, this quantity grows with the number of samples according to a power law whose exponent is determined by $d$, so PHD estimates $d$ by fitting this growth rate across subsamples of increasing size. Because it depends on the connectivity of the entire point cloud rather than on individual neighborhoods, PHD complements the two local estimators. It is also differentiable with respect to the sample locations, which allows it to be used as a training objective. Together, the three estimators span two families: local neighborhood statistics (2-NN, MLE) and global topological geometry (PHD).

Table~\ref{table:id_estimators} reports $\idh$ under all three estimators along the weight-decay sweeps for Halfcheetah and Cheetah-run (20K training samples each), together with the estimator-independent quantities (test MSE, NRC1, and, for Cheetah-run, policy return) and each estimator's estimate of $\idy$. Two observations show that our conclusions are not sensitive to the choice of estimator.

First, the three estimators agree closely in absolute value. Relative to 2-NN, the mean absolute deviation in $\idh$ along a sweep is 0.25--0.50 and the largest single deviation is 1.13, small compared with the range of roughly 4.5--6.3 over which $\idh$ varies along each sweep. The estimate of $\idy$ shifts by at most 0.29. Since test MSE, return, and NRC1 do not depend on the estimator, estimators that agree this closely trace out the same relationships between $\idh$, NRC1, and performance that underlie Sections~\ref{sec:nrc_geometry} and~\ref{sec:generalization}.

Second, and more fundamentally, our conclusions rely on the ordering of models along $\idh$ and on where $\idh$ crosses $\idy$, rather than on absolute ID values. The ordering is preserved across estimators, apart from swaps among near-tied non-collapsed models at the smallest weight decays. The regime boundary is also preserved: on Halfcheetah, all three estimators place the transition from $\idh > \idy$ to $\idh < \idy$ between $\lambda_{\text{WD}} = 10^{-3}$ and $5 \times 10^{-3}$; on Cheetah-run, 2-NN and MLE place it between $5 \times 10^{-4}$ and $10^{-3}$, while PHD places it one step earlier because the $\lambda_{\text{WD}} = 5 \times 10^{-4}$ model falls marginally below $\idy$ (6.00 vs.\ 6.03). In particular, the best-performing Cheetah-run model ($\lambda_{\text{WD}} = 10^{-3}$; lowest test MSE and highest return) lies next to the $\idh \approx \idy$ transition under every estimator.

\begin{table}[htbp]
\caption{Intrinsic dimension estimated by 2-NN, MLE, and PHD along weight-decay sweeps. Columns correspond to weight decay $\lambda_{\text{WD}}$. Each estimator row reports $\idh$ for every model. The last column reports the corresponding estimate of $\idy$, computed on the 20K training set, which is hence slightly different from the 80K value reported in Section~\ref{sec:datasets} for Cheetah-run. Test MSE, return, and NRC1 do not depend on the estimator. Bold numbers mark the best value of each performance metric across the sweep.}
\label{table:id_estimators}
\centering
\small
\vspace{4pt}
\textbf{Halfcheetah (20K)}\\[3pt]
\begin{tabular}{lccccccc c}
\toprule
 & \multicolumn{7}{c}{$\lambda_{\text{WD}}$} & \\
\cmidrule(lr){2-8}
 & 0 & 1E-5 & 1E-4 & 5E-4 & 1E-3 & 5E-3 & 1E-2 & $\idy$ \\
\midrule
Test MSE & 0.281 & 0.271 & \textbf{0.265} & 0.293 & 0.352 & 0.574 & 0.703 & -- \\
NRC1 & 0.251 & 0.235 & 0.166 & 0.059 & 0.034 & 0.005 & 0.001 & -- \\
\midrule
2-NN & 7.89 & 7.97 & 7.86 & 6.96 & 6.17 & 3.89 & 3.50 & 5.27 \\
MLE & 7.78 & 7.77 & 7.84 & 6.75 & 5.98 & 4.40 & 4.02 & 5.40 \\
PHD & 7.28 & 7.23 & 7.29 & 6.28 & 5.61 & 4.06 & 3.69 & 5.18 \\
\bottomrule
\end{tabular}

\vspace{8pt}
\textbf{Cheetah-run (20K)}\\[3pt]
\begin{tabular}{lcccccccc c}
\toprule
 & \multicolumn{8}{c}{$\lambda_{\text{WD}}$} & \\
\cmidrule(lr){2-9}
 & 0 & 1E-5 & 5E-5 & 1E-4 & 5E-4 & 1E-3 & 5E-3 & 1E-2 & $\idy$ \\
\midrule
Test MSE & 3.352 & 3.241 & 3.093 & 3.001 & 2.377 & \textbf{2.210} & 2.435 & 2.725 & -- \\
Return & 28.9 & 40.1 & 30.8 & 33.0 & 34.3 & \textbf{55.4} & 27.8 & 14.4 & -- \\
NRC1 & 0.645 & 0.665 & 0.673 & 0.648 & 0.021 & 0.023 & 0.014 & 0.007 & -- \\
\midrule
2-NN & 9.79 & 9.85 & 10.04 & 10.00 & 6.43 & 5.45 & 4.79 & 3.76 & 6.14 \\
MLE & 10.32 & 10.69 & 10.96 & 11.13 & 6.47 & 5.44 & 4.82 & 4.06 & 6.43 \\
PHD & 9.10 & 9.40 & 9.67 & 9.67 & 6.00 & 4.94 & 4.31 & 3.79 & 6.03 \\
\bottomrule
\end{tabular}
\end{table}

\subsection{Intrinsic Dimension versus Spectral Metrics} \label{appx:spectral_metrics}
Metrics computed from the eigenvalue spectrum of the feature covariance are popular proxies for representation quality. In deep reinforcement learning, the effective rank of the features is widely used to diagnose implicit under-parameterization and loss of plasticity~\citep{kumar2021implicit, lyle2022understanding}. In self-supervised learning, RankMe~\citep{garrido2023rankme}, an entropy-based effective rank of the feature matrix, and the power-law decay of the eigenspectrum~\citep{ghosh2022investigating} are used to select encoders with diverse features. NRC1 and the explained variance ratio (EVR) of the leading principal components belong to the same family. In neural regression collapse, however, such metrics saturate: \citet[Figure~3]{andriopoulos2024prevalence} show that, after a short period of training, the variance of the last-layer features concentrates on the first $n$ principal components, with little or no variance in the remaining ones. From then on, a spectral metric can change only through the relative magnitudes of the top-$n$ eigenvalues and is blind to the nonlinear structure of the features within the collapsed subspace. Spectral metrics therefore indicate \emph{whether} collapse has occurred, but not the geometry \emph{within} the collapsed subspace, which, according to our analysis, is where generalization is decided.

This limitation is intrinsic to any metric that depends on the data only through its covariance. Consider $X$ uniformly distributed on the unit circle in $\mathbb{R}^2$ and $Y \sim \mathcal{N}(0, \tfrac{1}{2}\bI_2)$. Writing $X = (\cos\theta, \sin\theta)$ with $\theta$ uniform on $[0, 2\pi)$, both distributions are centered and satisfy $\mathbb{E}[XX^\top] = \tfrac{1}{2}\bI_2 = \mathbb{E}[YY^\top]$. Their spectra are identical, so every covariance-based metric, including effective rank, assigns them the same value (for instance, RankMe equals 2 in both cases). Their intrinsic dimensions, however, differ: $\mathrm{ID}(X) = 1$ and $\mathrm{ID}(Y) = 2$. This is precisely the configuration of collapsed regression features, which lie in a low-dimensional linear subspace while occupying a curved, lower-dimensional manifold within it (Figures~\ref{fig:nrc_geometry} and~\ref{fig:nonlinear_manifold}).

Table~\ref{table:rankme} confirms this empirically by additionally comparing RankMe with $\idh$ along the weight-decay sweep on Cheetah-run. RankMe saturates exactly where NRC1 does: from $\lambda_{\text{WD}} = 5 \times 10^{-4}$ onward, it stays within 3.16--3.63, whereas $\idh$ continues to decrease from 6.43 to 3.76, crossing $\idy = 6.14$ into the over-compressed regime. Yet this saturated range is where model quality is decided: test MSE traces its U-shape and return varies from 14.4 to 55.4. Selecting the model with the highest rank, as RankMe prescribes, would favor the non-collapsed models ($\lambda_{\text{WD}} \le 10^{-4}$), all of which have higher test MSE and lower return than the best model. In contrast, the two models closest to $\idy$ attain the two lowest test MSEs and include the best-performing policy.

\begin{table}[htbp]
\caption{Spectral rank (RankMe) versus intrinsic dimension along the weight-decay sweep on Cheetah-run (20K). Columns correspond to weight decay $\lambda_{\text{WD}}$. The RankMe and ID rows are computed on the last-layer features of each model and the last column instead evaluates both metrics on the targets $Y$. Bold numbers mark the best value of each performance metric across the sweep.}
\label{table:rankme}
\centering
\small
\vspace{4pt}
\begin{tabular}{lcccccccc c}
\toprule
 & \multicolumn{8}{c}{$\lambda_{\text{WD}}$} & \\
\cmidrule(lr){2-9}
 & 0 & 1E-5 & 5E-5 & 1E-4 & 5E-4 & 1E-3 & 5E-3 & 1E-2 & $Y$ \\
\midrule
Test MSE & 3.352 & 3.241 & 3.093 & 3.001 & 2.377 & \textbf{2.210} & 2.435 & 2.725 & -- \\
Return & 28.9 & 40.1 & 30.8 & 33.0 & 34.3 & \textbf{55.4} & 27.8 & 14.4 & -- \\
NRC1 & 0.645 & 0.665 & 0.673 & 0.648 & 0.021 & 0.023 & 0.014 & 0.007 & -- \\
\midrule
RankMe & 73.10 & 76.76 & 77.25 & 69.89 & 3.54 & 3.63 & 3.57 & 3.16 & 3.53 \\
ID (2-NN) & 9.79 & 9.85 & 10.04 & 10.00 & 6.43 & 5.45 & 4.79 & 3.76 & 6.14 \\
\bottomrule
\end{tabular}
\end{table}

\section{Evolution of Intrinsic Dimension during Training}
\label{appx: id_evolution}
\begin{figure}[htb]
    \centering
    \includegraphics[width=0.8\linewidth]{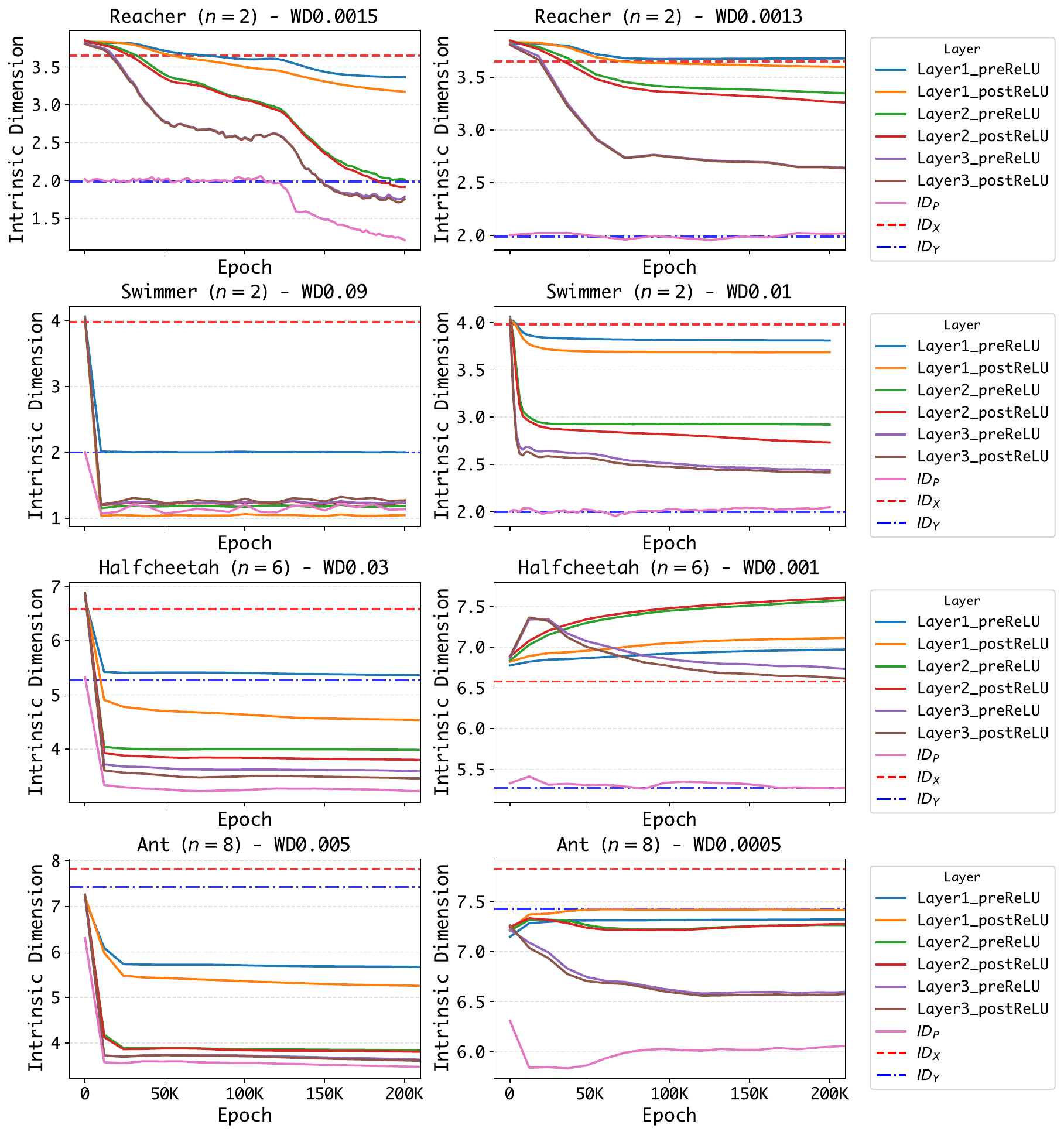}
    \caption{Intrinsic dimension of input, output, and hidden layers over training epochs for a collapsed (left) and a non-collapsed model (right) for MuJoCo datasets. Each subfigure shows the evolution of intrinsic dimension across layers with red dashed, blue dash-dotted, and pink lines denoting the intrinsic dimension of inputs, targets, and predicted outputs, respectively.}
    \label{fig_appx:id_evolution}
\end{figure}

To further understand the behavior of collapsed models and their counterparts, we track the evolution of the intrinsic dimension throughout training and provide insights. Figure \ref{fig_appx:id_evolution} provides illustrative examples for a collapsed and a non-collapsed model:

\begin{itemize}
    \item For both the collapsed and non-collapsed models, the intrinsic dimension of the last-layer features generally decreases over training and stabilizes at convergence. 
    \item For the collapsed model, the deeper the layer in the network, the lower the intrinsic dimension at the end of training. ReLU activations cause a mild reduction in intrinsic dimension in comparison with the reduction in intrinsic dimension between consecutive layers (ignoring ReLU). Notably, the final intrinsic dimension of the output layer, which gives the actual vector-valued predictions, can be significantly lower than $\idh$.
    \item For non-collapsed models, we usually see --- but not always --- $\idh$ decrease monotonically as we move from shallow to deep layers. Furthermore, we observe that during training, the intrinsic dimension of the output layer hugs the intrinsic dimension of the targets. Thus, tracking the intrinsic dimension of the output layer provides yet another criterion for discriminating between collapsed and non-collapsed models; see Appendix~\ref{appx:output_layer}.  
\end{itemize}

\newpage
\section{Intrinsic Dimension and Output layer}
\label{appx:output_layer}
We consider here the intrinsic dimension of the outputs (equivalently, the final predictions), $\idp$.
We will see that here too the relationship between intrinsic dimension and generalization exhibits key differences between classification and regression. 

\begin{figure}[htb]
    \centering
    \includegraphics[width=\linewidth]{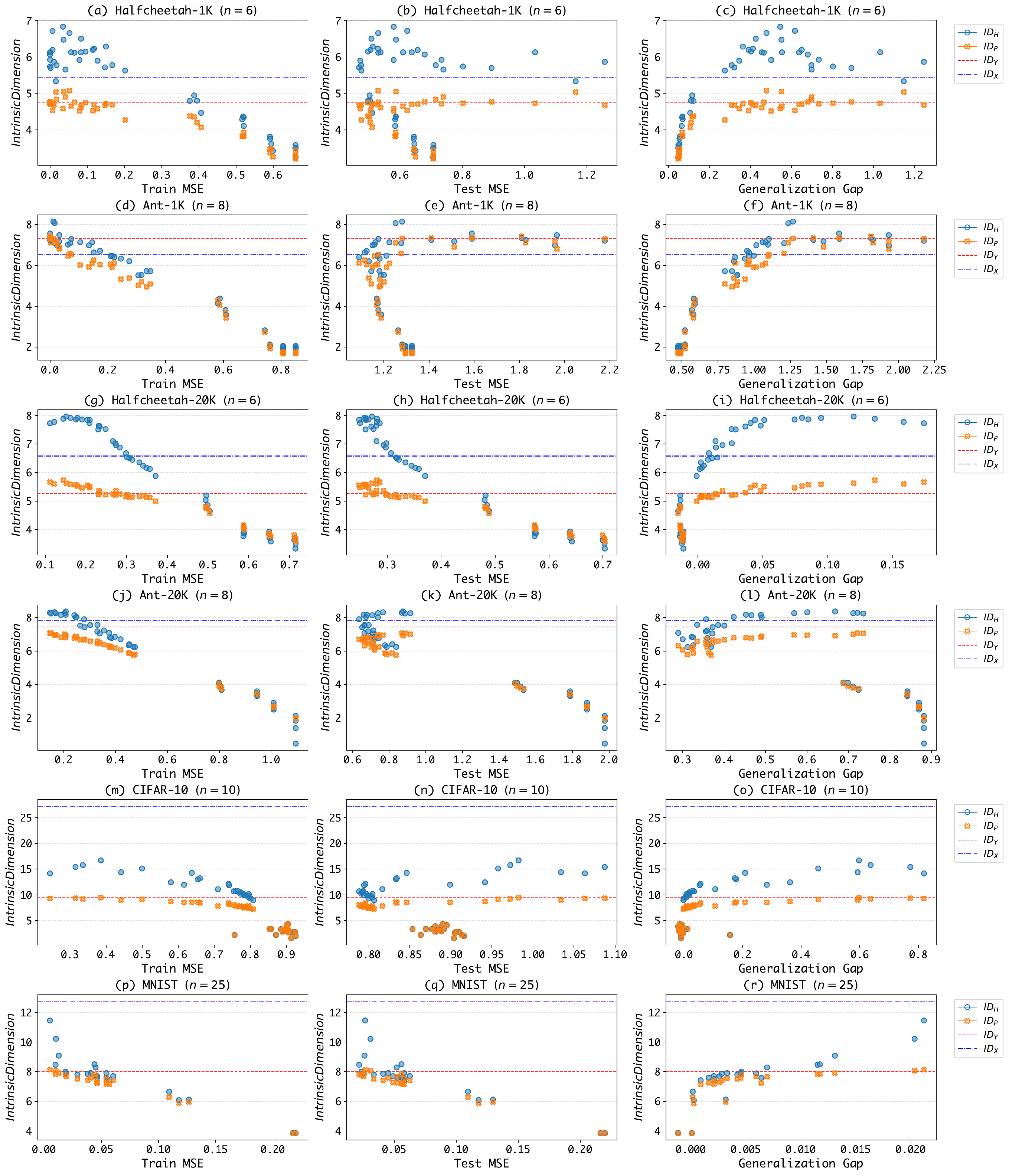}
    \caption{Comparison between $\idh$ and $\idp$ for Halfcheetah, Ant, CIFAR-10, and MNIST datasets}
    \label{fig:6x3_idp1}
\end{figure}

With respect to the output layer, a structural constraint arises from the classification setting. Specifically, the intrinsic dimension of the output layer necessarily satisfies
\[
\log_2 C\le \idp\le C,
\]
where $C$ is the number of classes. Empirical results consistently show $\idp$ equals the lower bound of this inequality if the model generalizes well. We refer the reader to the discussion in Section 3.1 of \citet{ansuini2019intrinsic}. Conversely, saturation of the upper bound, i.e., $\idp\approx C$, is associated with poor generalization performance, suggesting that maximal output layer dimensionality corresponds to overfitting in classification tasks, see Section 3.5 in \citet{ansuini2019intrinsic}.

\begin{figure}[htb]
    \centering
    \includegraphics[width=\linewidth]{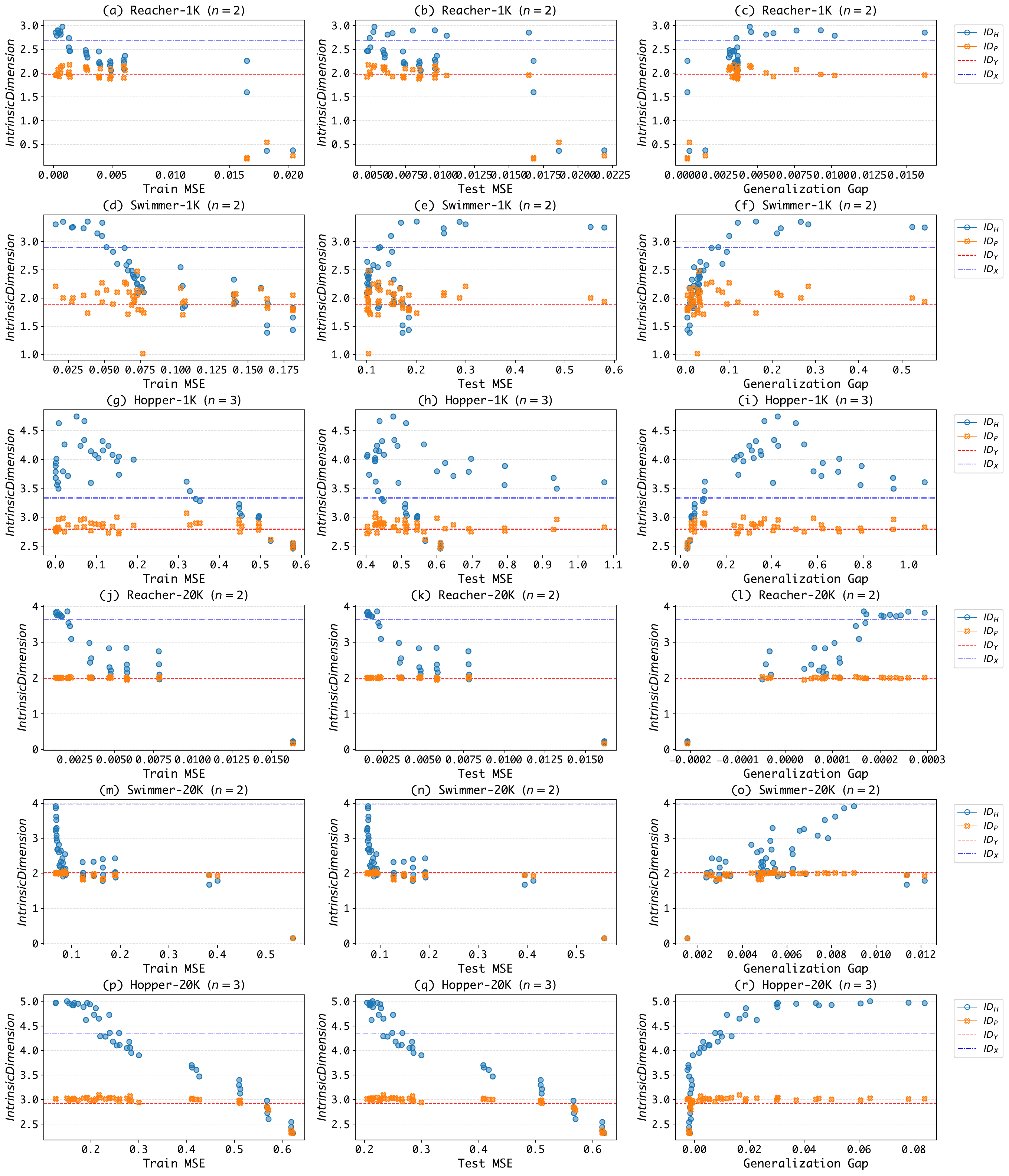}
    \caption{Comparison between $\idh$ and $\idp$ for Reacher, Swimmer and Hopper datasets}
    \label{fig:6x3_idp2}
\end{figure}

In contrast, for neural multivariate regression, the structure of the output leads to the trivial bound
\[
1\le \idp\le n,
\]
where $n$ is the number of output variates. Interestingly, our empirical findings reveal a departure from the classification setting. As shown in the middle column in Figures~\ref{fig:6x3_idp1}-\ref{fig:6x3_idp2}, when the test MSE is low, the intrinsic dimension of the output layer, $\idp$ satisfies $\idp \approx \idy$, which can be close to $n$,  
saturating the upper bound of the inequality above.
Notably, unlike in classification, this saturation is associated with improved test performance. By contrast, when $\idp$ falls below $\idy$, test MSE performance deteriorates, see Figures~\ref{fig:6x3_idp1}-\ref{fig:6x3_idp2}.

\clearpage
\section{Additional Experiments on Generalization} 
\label{appx:generalization}

\begin{figure}[htb]
    \centering
    \includegraphics[width=\linewidth]{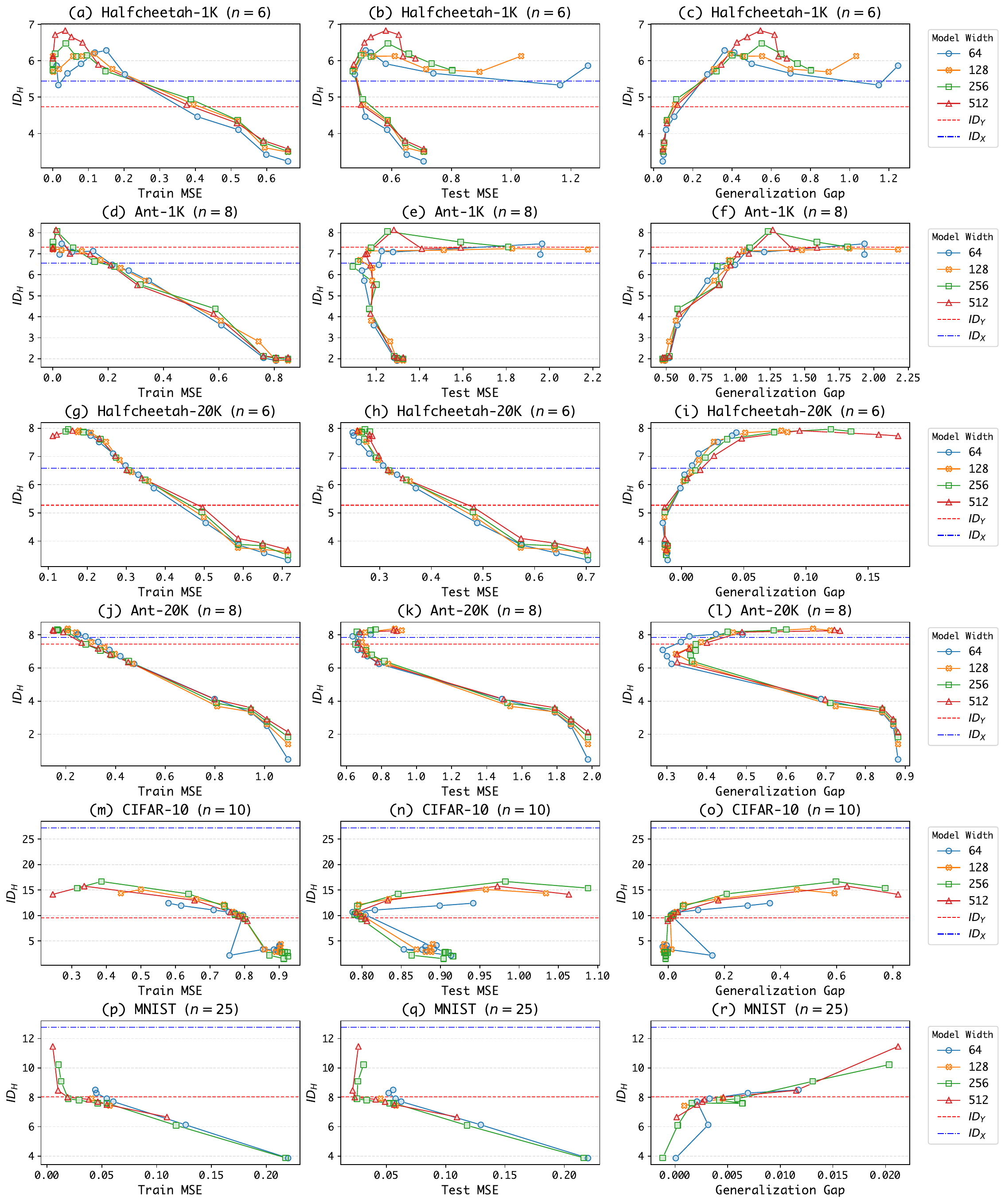}
    \caption{Generalization ability and Intrinsic Dimension for Halfcheetah, Ant, CIFAR-10, and MNIST datasets.}
    \label{fig_appx:generalization_appx_1}
\end{figure}

\begin{figure}[htb]
    \centering
    \includegraphics[width=\linewidth]{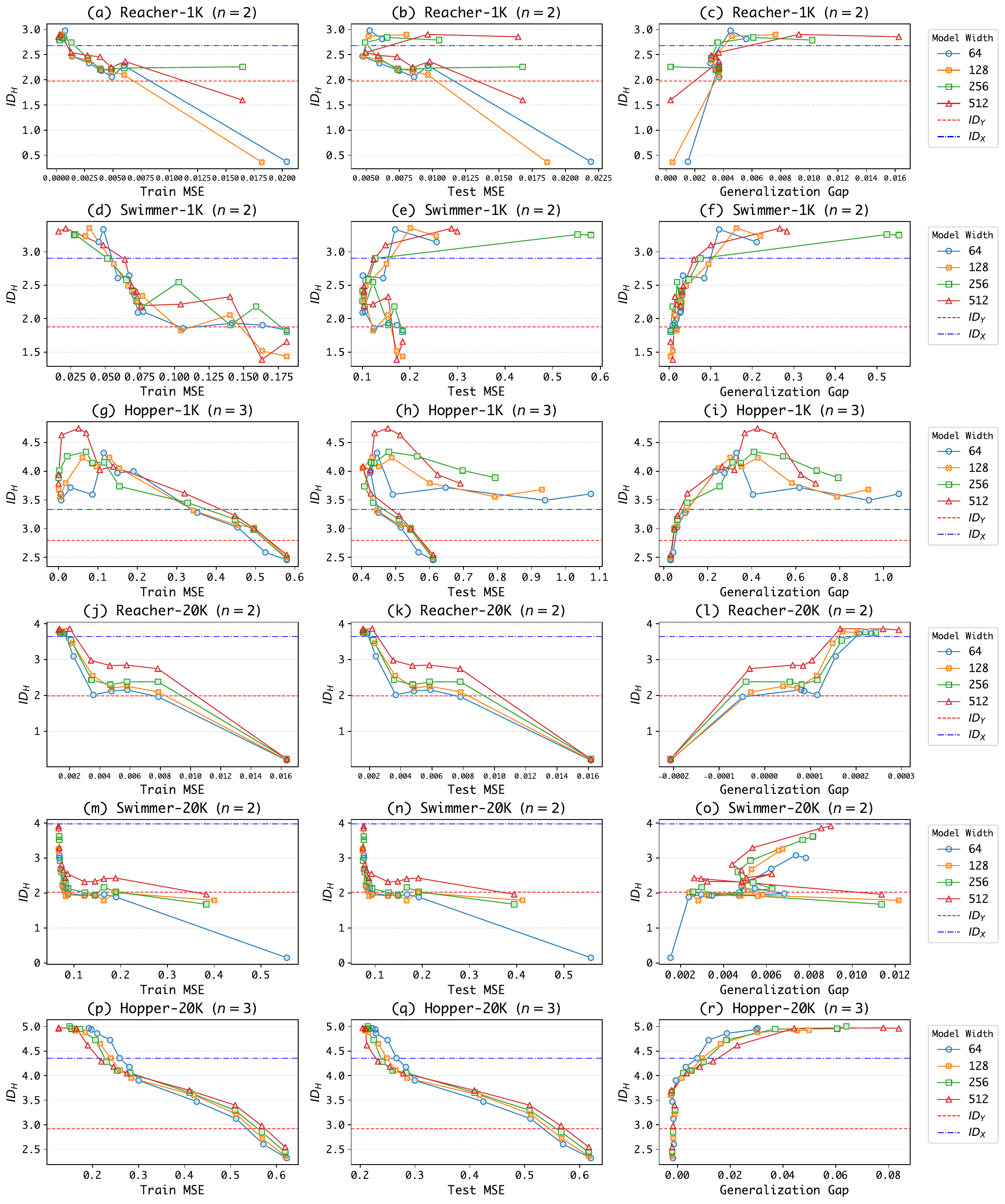}
    \caption{Generalization ability and Intrinsic Dimension for Reacher, Swimmer, and Hopper datasets.}
    \label{fig_appx:generalization_appx_2}
\end{figure}

This section lists additional results that complement the experiments in the main body for all considered datasets. Figure~\ref{fig_appx:generalization_appx_1} and ~\ref{fig_appx:generalization_appx_2} shows how generalization power correlates with $\idh$. Our key takeaways are summarized in Table~\ref{table:gen}.

\clearpage 
\section{How does regularization affect NRC and Intrinsic Dimension?}
\label{appx:nrc1_regularization}
\subsection{Weight Decay} \label{appx: nrc1_wd}
Weight decay is a canonical and widely adopted model regularization technique for preventing large models from overfitting to data. Real-world applications include but are not limited to (1) regularizing transformer backbones for large language models \citep{wolf-etal-2020-transformers} and robotic generalist policies \citep{chen2021decision}; (2) participating, by default, in the common Pytorch implementation of AdamW optimizer \citep{loshchilov2018adamw, pytorchAdamW} with $\texttt{weight\_decay=0.01}$; (3) improving sample efficiency of online reinforcement learning algorithms \citep{liu2021regularization, li2023efficient}; (4) and facilitating research in model plasticity in deep learning \citep{lyle2023understanding, nauman2024overestimation, ceron2024pruned}.

\begin{figure}[htpb]
    \centering
    \includegraphics[width=\linewidth]{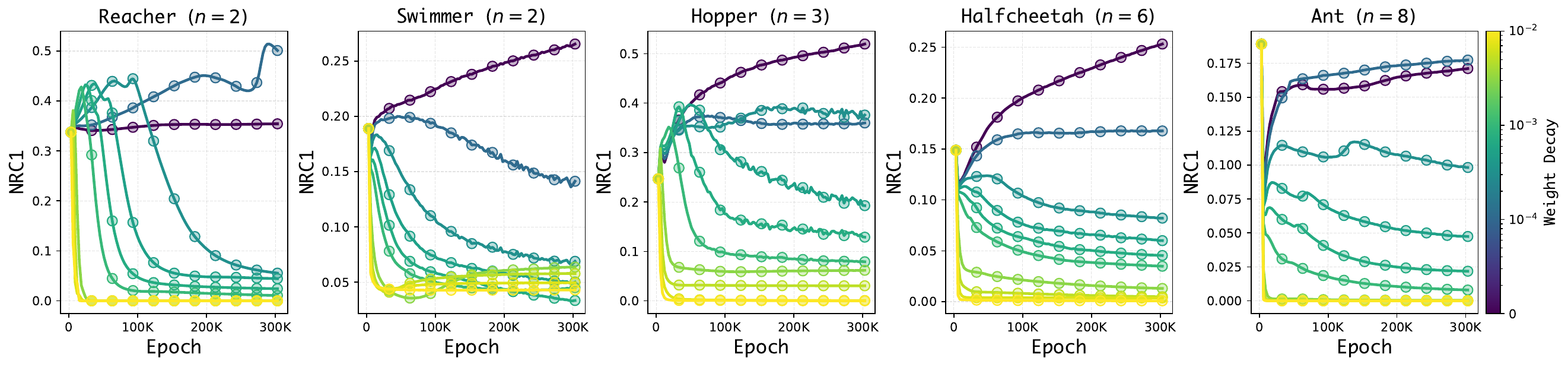}
    \caption{NRC1 decreases as weight decay becomes stronger, leading to model collapse.}
    \label{fig_appx:nrc1_wd}
\end{figure}

Figure \ref{fig_appx:nrc1_wd} investigates NRC1 for values of the weight decay parameter $\lambda_{\text{WD}}$. We see that when $\lambda_{\text{WD}}$ is zero or small, there is no neural regression collapse; but if we increase the weight decay, the NRC1 geometric structure quickly emerges during training. This matches the empirical observation in \citet{andriopoulos2024prevalence}.

\subsection{Dropout Regularization}
\begin{figure}[ht]
    \centering

    \subfloat[NRC1 - $\idh$]{%
        \includegraphics[width=0.65\linewidth]{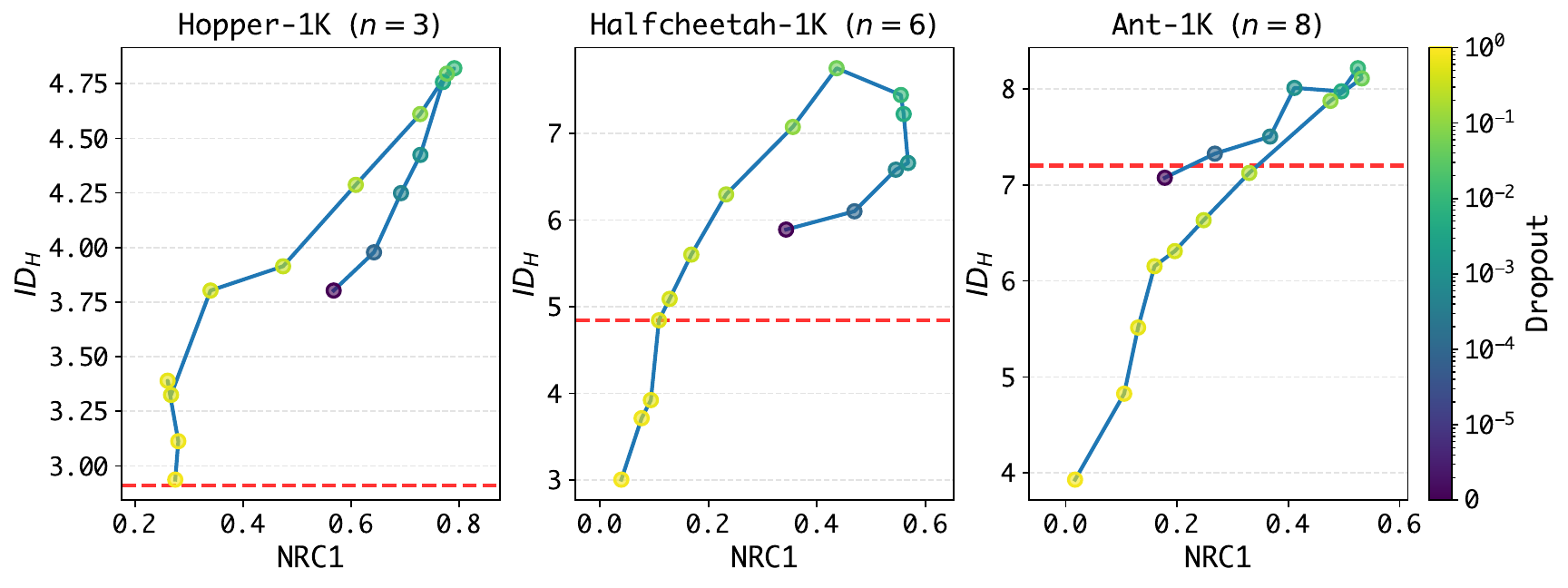}
    }\\[1em]

    \subfloat[Test MSE - $\idh$]{%
        \includegraphics[width=0.65\linewidth]{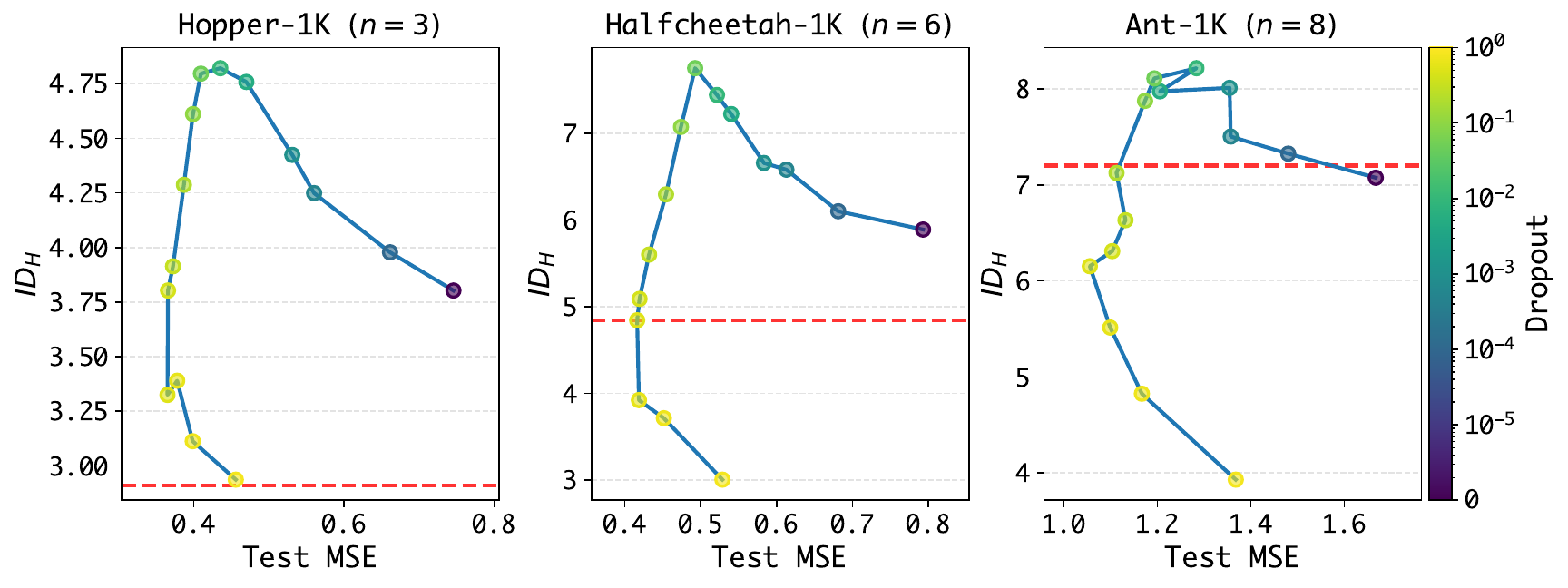}
    }

    \caption{Relationship between $\idh$ and NRC1 and Test MSE for Hopper-1K, Halfcheetah-1K, and Ant-1K datasets, when applying model dropout regularization. The horizontal red dashed line represents $\idy$.}
    \label{fig:dropout_1k}
\end{figure}

\begin{figure}[ht]
    \centering

    \subfloat[NRC1 - $\idh$]{%
        \includegraphics[width=0.65\linewidth]{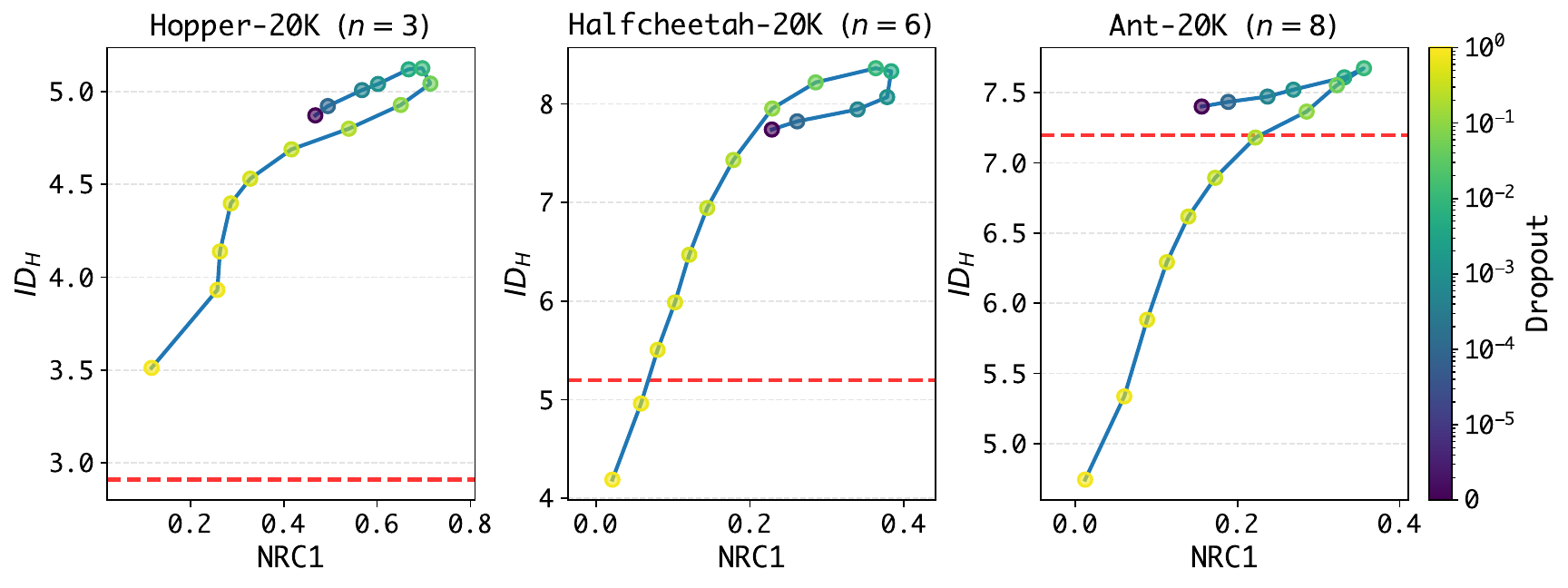}
    }\\[1em]

    \subfloat[Test MSE - $\idh$]{%
        \includegraphics[width=0.65\linewidth]{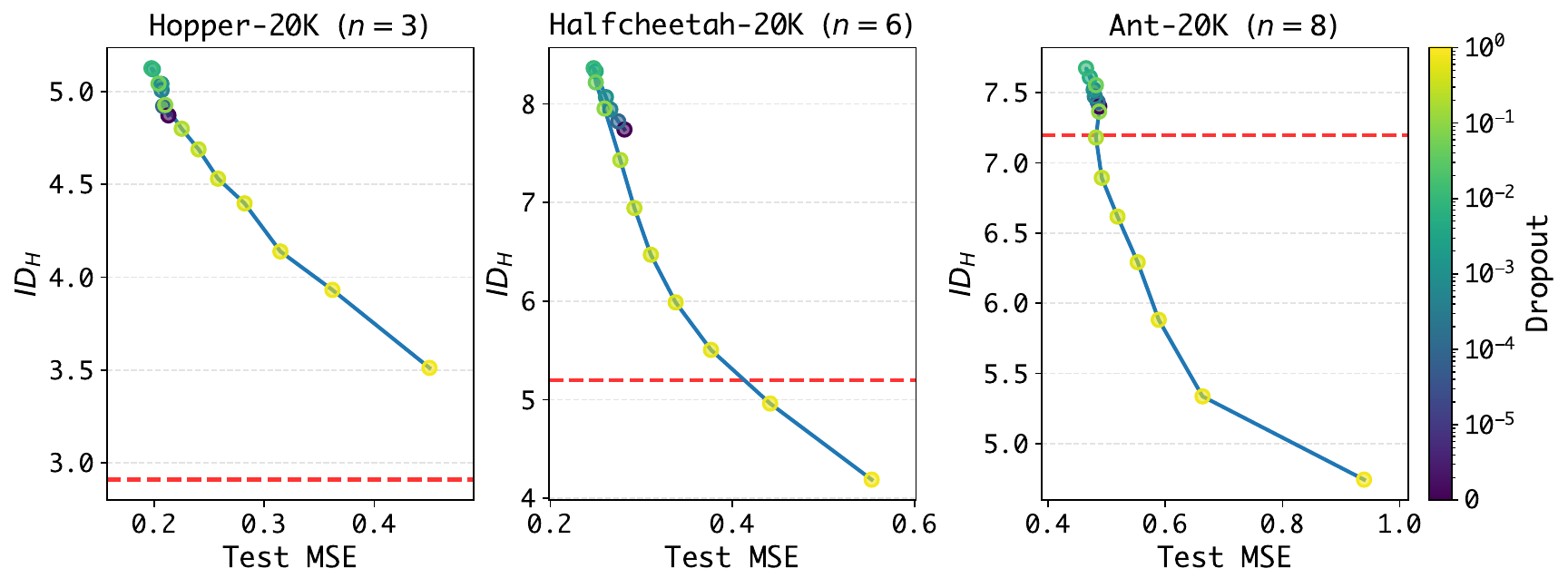}
    }

    \caption{Varying model dropout regularization, relationship between $\idh$ and NRC1 and Test MSE for Hopper-20K, Halfcheetah-20K, and Ant-20K datasets. The horizontal red dashed line represents $\idy$.}
    \label{fig:dropout_20k}
\end{figure}

Modern implementations, across deep learning domains, continue to rely on dropout regularization \citep{srivastava14adropout} to mitigate overfitting, underscoring its persistent role in practical training, such as computer vision \citep{dosovitskiy2021vit}, NLP \citep{devlin2019bert, wolf-etal-2020-transformers}, and reinforcement learning \citep{hiraoka2022droq}. We empirically analyze how dropout regularization influences neural regression collapse by varying its strength from a large range ($\in \{0, 0.0001, 0.0005, 0.001, 0.005, 0.01, 0.05, 0.1, 0.2, 0.3, 0.4, 0.5, 0.6, 0.7, 0.8\}$). \emph{No} weight decay is applied. In this section, datasets include Hopper, Halfcheetah, and Ant with two sizes, 1K (Figure~\ref{fig:dropout_1k}) and 20K (Figure~\ref{fig:dropout_20k}). The horizontal red dashed line represents $\idy$.

Figure~\ref{fig:dropout_1k}(a) and Figure~\ref{fig:dropout_20k}(a) show the relationship between $\idh$ and NRC1. We first confirm the same conclusion as made in Section 5, despite the new regularization. $\idh$ provides a more refined geometric structure than NRC1. Collapsed models with near-zero NRC1 values have varying $\idh$ below or in the vicinity of $\idy < n$, while non-collapsed models with non-trivial NRC1 maintain their $\idh$ to be above $\idy$ and to be positively correlated with $NRC1$. Interestingly, dropout regularization differs from weight decay in that mild dropout (e.g., $\leq 0.01$) can effectively prevent models from collapse by increasing both NRC1 and $\idh$. This observation sheds light on a geometric interpretation of the effectiveness of mild dropout in reinforcement learning as proposed by \citet{hiraoka2022droq}.

Figure~\ref{fig:dropout_1k}(b) and Figure~\ref{fig:dropout_20k}(b) show the relationship between $\idh$ and test MSE. The results again verify the three regimes discussed in Section 6. For both data sizes, models over-compress features when $\idh < \idy$ and thus lead to increasing test MSE (and thus poor generalization). Then, for small datasets with 1K samples, $\idh \approx \idy$ identifies the sweet spot where test MSE tends to be the lowest and exhibits the `U-shape' plots. Note that models trained with Hopper datasets have not collapsed yet, so they only exhibit the upper part of the `U-shape'. Finally, with more samples, e.g., 20K, $\idh \gg \idy$ achieves the best generalization with under-compressed models. 

In summary, dropout regularization offers an alternative approach to adjusting the degree of model collapse, while all conclusions drawn from the main body remain intact and inclusive. In addition, mild dropout regularization is more effective than weight decay regularization in increasing NRC1 and $\idh$ metrics for the under-compressed regime.

\subsection{Model Depth}
\begin{figure}[h]
    \centering
    \includegraphics[width=0.6\linewidth]{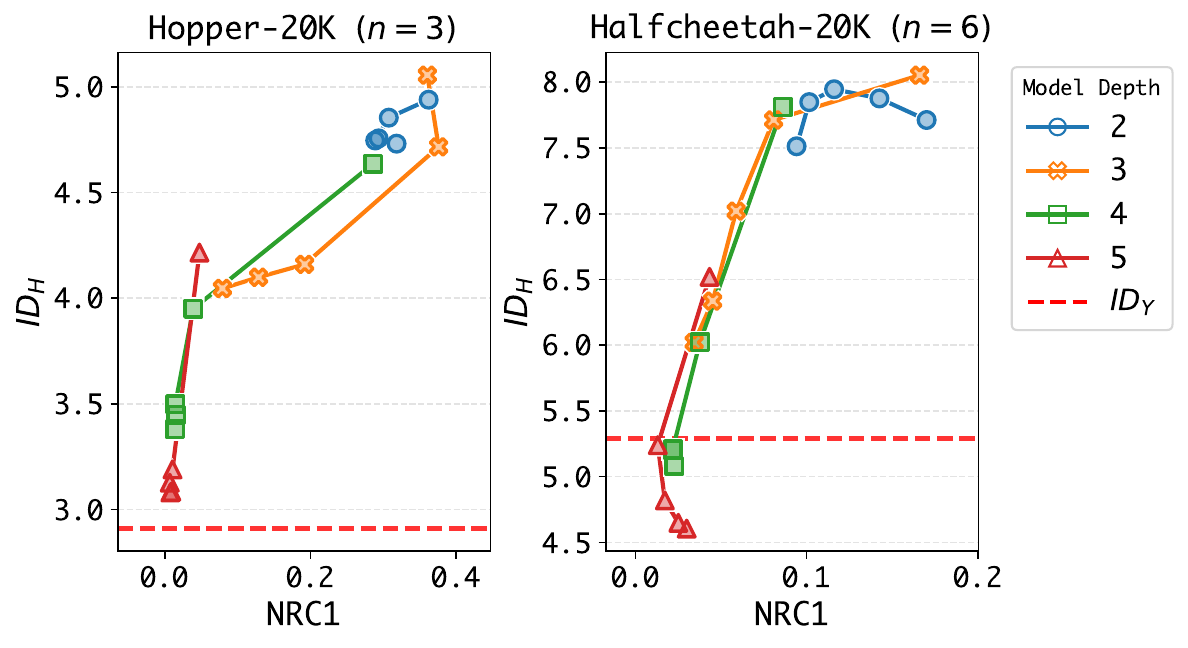}
    \caption{Relationship between $\idh$ and NRC1 for Hopper-20K and Halfcheetah-20K datasets, when varying model depth. All models have a hidden size of 256. The horizontal red dashed line represents $\idy$.}
    \label{fig:model_depth}
\end{figure}

With mild weight decay regularization, we find that increasing model depth leads to smaller NRC1 and $\idh$ and thus to more collapsed features. In Figure~\ref{fig:model_depth}, we examine the relationship between $\idh$ and NRC1 for Hopper-20K and Halfcheetah-20K datasets. We fix the model width to be 256 and vary the model depth ($\in \{2, 3, 4, 5\}$). For each model depth, we show five mild weight decay values: 0.0001, 0.0003, 0.0005, 0.0007, 0.001. The figure shows that increasing model depth gradually pushes points to the region on the bottom left, where both NRC1 and $\idh$ are small. For example, Halfcheetah-20K with a depth of 5 can result in collapsed models with $\idh < \idy$.

\section{Empirical Analysis on More Challenging Tasks}
\label{appx:more_tasks}
We extend our empirical analysis to more challenging tasks with varying data sizes, increased intrinsic dimensions, and visual inputs.

\begin{table}[h!]

\caption{Overview of additional datasets employed in this section.}
\begin{center}

\scalebox{0.75}{
\begin{tabular}{ccccccc}
\toprule
\textbf{Dataset} & \textbf{Data Size} & \textbf{Input Type} &\textbf{Input Dim ($D$)} & \textbf{Input ID ($\idx$)} & \textbf{Target Dim} ($n$) & \textbf{Target ID ($\idy$)}\\
\midrule
Humanoid & 500,000 & raw state & 348 & 11.02 & 17 & 9.85\\

Relocate & 500,000 & raw state & 39 & 6.90 & 30 & 19.82 \\

\midrule
Cheetah\_run & 80,000 & RGB image & $84\times84\times9$\footnotemark & 8.96 & 6 & 6.00\\

Humanoid\_walk & 100,000 & RGB image & $84\times84\times9$ & 9.41 & 21 & 14.84\\
    
\bottomrule
\end{tabular}
}

\end{center}
\label{table:dataset_appx}
\end{table} 

\footnotetext{Frame stack is commonly applied for visual control tasks. A single observation is of shape $84\times84\times3$. A frame stack of 3 is used in our experiments, resulting in visual inputs of dimension $84\times84\times9$.}

\paragraph{Humanoid (MuJoCo locomotion)}
The Humanoid dataset \citep{minari} is generated from the MuJoCo physics simulator, introduced in the main body and Appendix~\ref{appx: mujoco}. Each state consists of high-dimensional proprioceptive information, and the corresponding targets are the expert control torques applied at each joint. The goal is to enable the humanoid to run forward stably while maintaining balance, which is substantially more difficult than all previously considered MuJoCo tasks due to its high degrees of freedom and complex contact dynamics. Among all MuJoCo environments, Humanoid is widely regarded as the most challenging.

\paragraph{Relocate (Adroit manipulation)}
The Relocate dataset \citep{fu2020d4rl} comes from the Adroit suite \citep{rajeswaran18adroit} of dexterous manipulation tasks. Adroit uses a simulated 24 degrees of freedom (24-DoF) robotic hand, combined with an arm of up to 6-DoF, with rich contact and articulation dynamics. In the Relocate task, the state includes joint positions, velocities, hand pose information, and kinematic information about the ball and target. The action corresponds to the joint torques for the 24 actuators and to the arm movement. The objective is to grasp a small object and relocate it to a specified target position. It is a long-horizon manipulation task requiring precise coordination and contact control. Among the four Adroit tasks in the D4RL benchmark \citep{fu2020d4rl}, Relocate is widely considered the most difficult due to its combination of dexterity, precision, and exploration complexity.

\paragraph{Cheetah\_run \& Humanoid\_walk (Visual continuous control)}
The Cheetah\_run and Humanoid\_walk datasets \citep{lu2023vd4rl} are visual control benchmarks constructed from demonstrations generated in the DeepMind Control Suite \citep{tassa2018deepmindcontrolsuite}. Inputs consist of raw image observations (e.g., 84×84x3 RGB frames), and targets correspond to the continuous control commands. In this section, the Cheetah\_run dataset consists of expert demonstrations, as is the case for all previous datasets, while the Humanoid\_walk dataset contains some noisy suboptimal behavior in addition to the expert demonstrations (`medium-expert dataset'). This adds difficulty in extracting a good policy by imitating the dataset's behavior. We use a CNN image encoder (consisting of 4 \texttt{conv2d} layers and ReLU activation), followed by a 3-layer MLP policy network.

Visual control is particularly interesting and challenging because each frame provides only partial information about the system state, effectively forming a Partially Observed Markov Decision Process (POMDP) \citep{yarats2022drq_v2}. As a result, the model must infer underlying transition dynamics and identify salient visual features directly from high-dimensional pixel inputs. Within the visual D4RL tasks \citep{lu2023vd4rl}, Humanoid\_walk is the most challenging due to the humanoid’s instability and high-dimensional dynamics, whereas Cheetah\_run is comparatively easier but still nontrivial.

\begin{figure}[ht]
    \centering

    \subfloat[NRC1 - $\idh$]{%
        \includegraphics[width=0.90\linewidth]{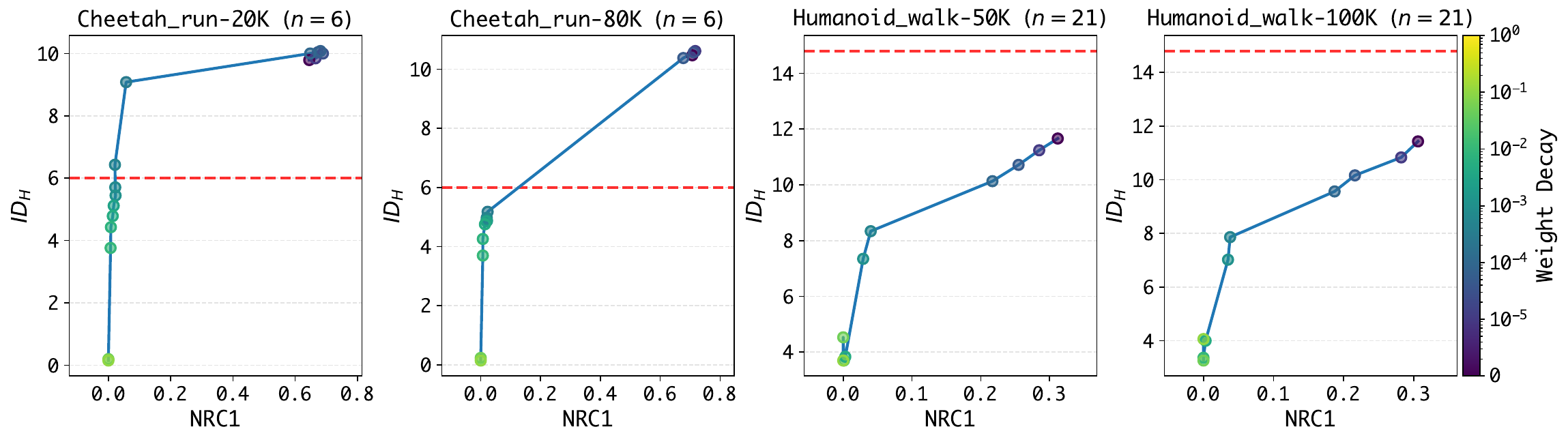}
    }\\[1em]

    \subfloat[Test MSE - $\idh$]{%
        \includegraphics[width=0.90\linewidth]{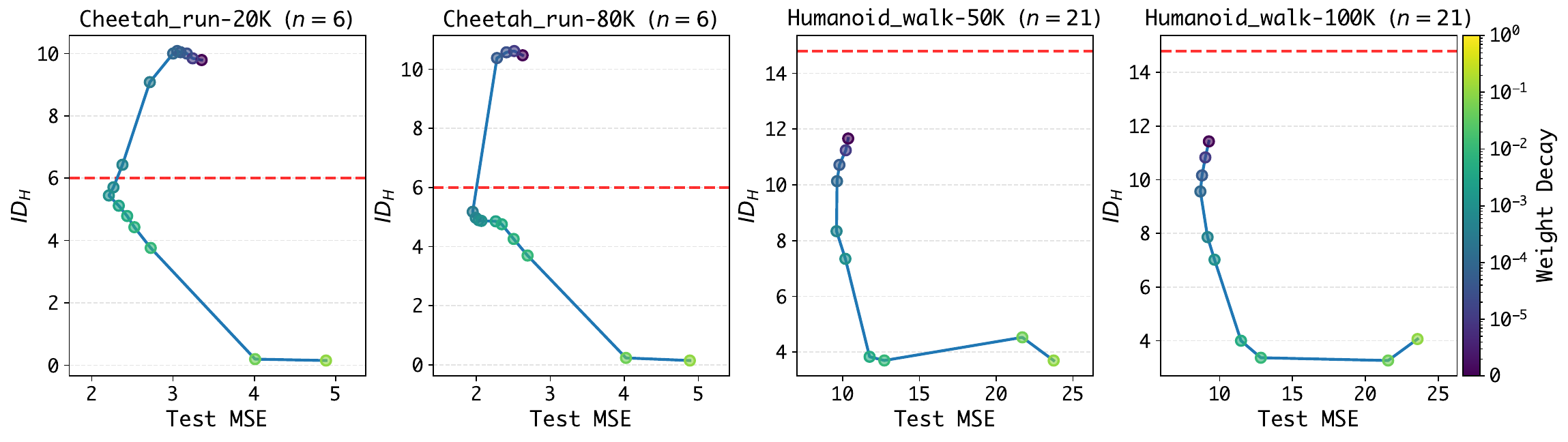}
    }

    \caption{Relationship between $\idh$ and NRC1 and Test MSE for Cheetah\_run-20K, Cheetah\_run-80K, Humanoid\_walk-50K and Humanoid\_walk-100K. The horizontal red dashed line represents $\idy$.}
    \label{fig:vd4rl_main}
\end{figure}

\subsection{Visual Control Tasks} 
Figure~\ref{fig:vd4rl_main}(a) shows the relationship between $\idh$ and NRC1 for the visual control datasets with varying sizes. Consistent with the conclusions in Section 5, $\idh$ provides a more refined geometric structure than NRC1. Collapsed models with small NRC1 values have varying $\idh$ below or in the vicinity of $\idy < n$, while non-collapsed models with non-trivial NRC1 maintain their $\idh$ to be above $\idy$ and to be positively correlated with NRC1. Notably, for the most challenging Humanoid\_walk task, which has a substantially higher $\idy$ due to its complex high-dimensional dynamics, models trained with \emph{zero} weight decay initially exhibit neural regression collapse with $\idh < \idy$. This explains the extremely large test MSE observed for Humanoid\_walk in Figure~\ref{fig:vd4rl_main}(b) and its negative correlation with $\idh$, which matches the over-compressed regime ($\idh < \idy$) summarized in Table~\ref{table:gen}. In comparison, the two Cheetah\_run datasets exhibit the `U-shape' in the relationship between $\idh$ and test MSE. This emphasizes the sweet spot with $\idh \approx \idy$, which approaches the lowest test MSE in noisy and challenging tasks.

\subsection{Relocate \& Humanoid}
\begin{figure}[ht]
    \centering

    \subfloat[NRC1 - $\idh$]{%
        \includegraphics[width=0.99\linewidth]{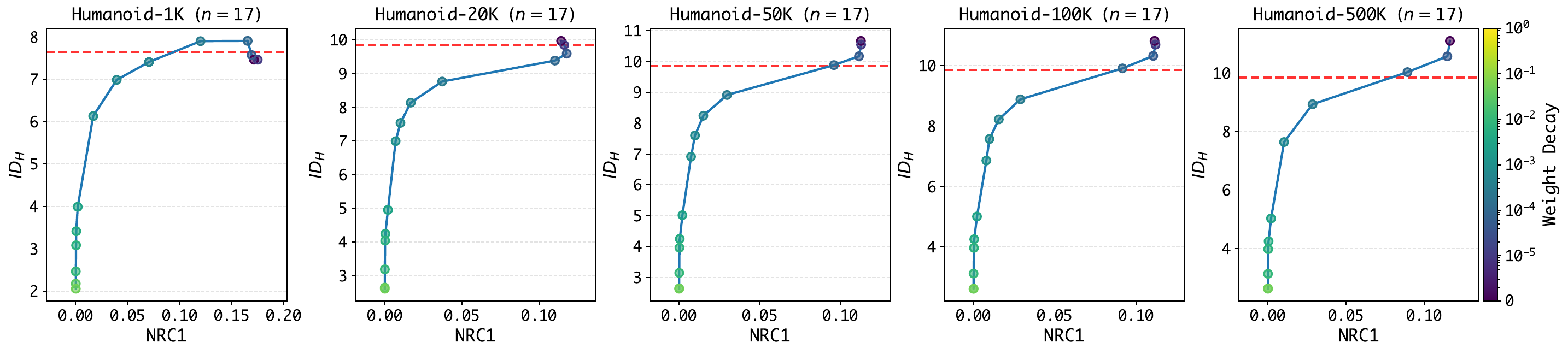}
    }\\[1em]

    \subfloat[Test MSE - $\idh$]{%
        \includegraphics[width=0.99\linewidth]{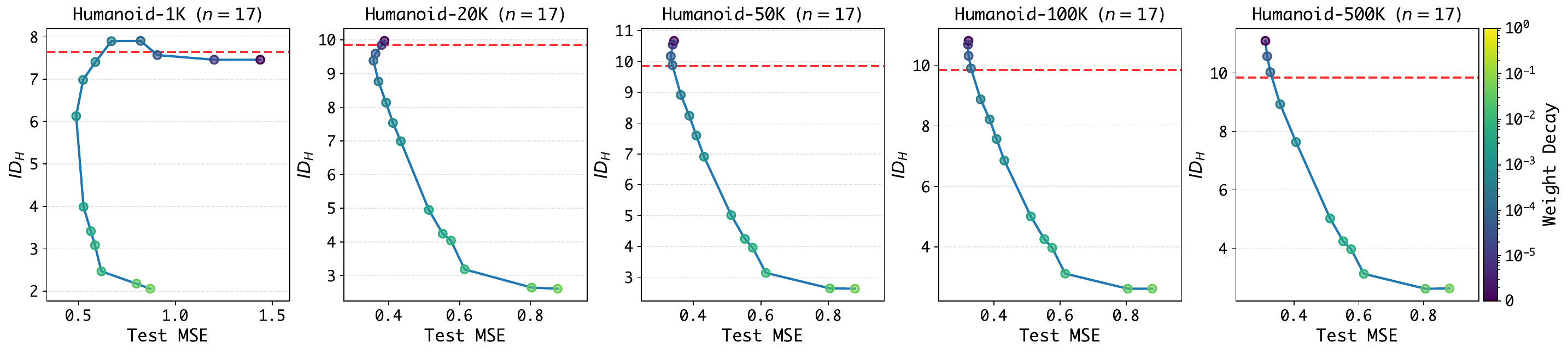}
    }

    \caption{Relationship between $\idh$ and NRC1 and Test MSE, for Humanoid locomotion task, when varying data size ($\in \{\text{1K}, \text{20K}, \text{50K}, \text{100K}, \text{500K}\}$). The red dashed line represents $\idy$.}
    \label{fig:size_humanoid}
\end{figure}

\paragraph{Humanoid} In Figure~\ref{fig:size_humanoid}, we show `NRC1 - $\idh$' and `Test MSE - $\idh$' plots for Humanoid datasets with varying sizes. In Figure~\ref{fig:size_humanoid}(a), we observe that $\idh$ provides a more refined geometric structure than NRC1. Collapsed models with near-zero NRC1 values have varying $\idh$ below or in the vicinity of $\idy < n$, while non-collapsed models with non-trivial NRC1 maintain their $\idh$ to be above $\idy$ and to be positively correlated with NRC1. Then, Figure~\ref{fig:size_humanoid}(b) again verifies the three regimes discussed in Section 6. For all data sizes, models over-compress features when $\idh < \idy$ and thus lead to increasing test MSE (and thus poor generalization). Then, for small datasets with 1-50K samples, $\idh \approx \idy$ identifies the sweet spot where test MSE tends to be the lowest and exhibits the `U-shape' plots. Finally, with more samples, e.g., 100-500K, $\idh > \idy$ achieves the best generalization with under-compressed models. 

\paragraph{Relocate} 
For the Relocate task with varying data sizes, which have the highest $\idy$ due to its complex task dynamics, models trained with \emph{zero} weight decay initially exhibit a severe collapse with both $\text{NRC1} \approx 0$ and $\idh \ll \idy$. Unsurprisingly, Test MSE remains large for all data sizes and decay values, and it also negatively correlates with $\idh$, uncovering the over-compressed regime.

\begin{figure}[ht]
    \centering

    \subfloat[NRC1 - $\idh$]{%
        \includegraphics[width=0.99\linewidth]{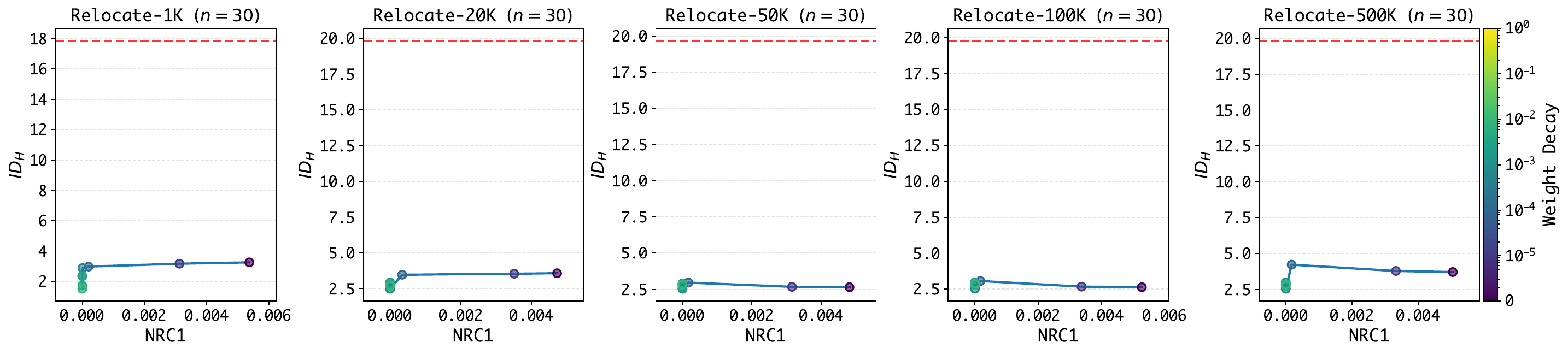}
    }\\[1em]

    \subfloat[Test MSE - $\idh$]{%
        \includegraphics[width=0.99\linewidth]{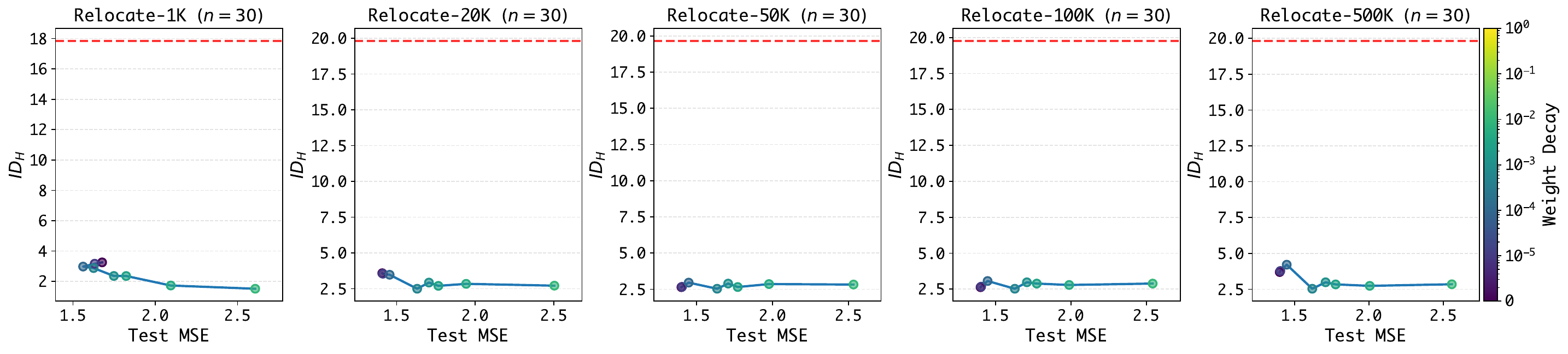}
    }

    \caption{Relationship between $\idh$ and NRC1 and Test MSE, for Relocate manipulation task, when varying data size ($\in \{\text{1K}, \text{20K}, \text{50K}, \text{100K}, \text{500K}\}$). The red dashed line represents $\idy$.}
    \label{fig:size_relocate}
\end{figure}

\clearpage
\section{Mathematical Analysis of Regression Collapse}
\label{appx:mathematical_analysis}

\subsection{Why Weight Decay Leads to Collapse: Analysis via the Unconstrained Feature Model}
\label{appx:ufm_collapse}

In this section, we provide a theoretical explanation for why weight decay causes neural regression collapse through the lens of the Unconstrained Feature Model (UFM). The UFM is a simplified mathematical abstraction that does not capture all aspects of practical neural networks, but provides insight into the geometric mechanisms by which regularization constrains learned representations.

The UFM abstracts the neural regression problem by treating the feature extractor $h_\theta(\cdot)$ as producing arbitrary feature vectors $\mathbf{h}_i \in \mathbb{R}^d$ for each input $\mathbf{x}_i$, collected into a feature matrix $H \in \mathbb{R}^{d \times M}$. The final prediction is obtained via a linear map $W \in \mathbb{R}^{n \times d}$ and bias $\mathbf{b} \in \mathbb{R}^n$, giving predictions $\hat{Y} = WH + \mathbf{b} \mathbf{1}^\top$. The UFM objective is:
\begin{equation}
\min_{W, H, \mathbf{b}} \frac{1}{2M} \|WH + \mathbf{b}\mathbf{1}^\top - Y\|_F^2 + \frac{\lambda_W}{2}\|W\|_F^2 + \frac{\lambda_H}{2}\|H\|_F^2
\end{equation}
The model is ``unconstrained'' because $H$ can be any matrix in $\mathbb{R}^{d \times M}$, unlike in actual neural networks, where $H$ is constrained by the input data and network architecture. In the UFM, we typically have $d \gg n$, mirroring the overparameterized regime where feature dimension greatly exceeds target dimension.

The optimal bias is $\mathbf{b}^* = \bar{\mathbf{y}}$ where $\bar{\mathbf{y}} = \frac{1}{M}\sum_{i=1}^M \mathbf{y}_i$. Defining the centered target matrix $\tilde{Y} = Y - \bar{\mathbf{y}}\mathbf{1}^\top \in \mathbb{R}^{n \times M}$, the problem reduces to:
\begin{equation}
\min_{W, H} \frac{1}{2M} \|WH - \tilde{Y}\|_F^2 + \frac{\lambda_W}{2}\|W\|_F^2 + \frac{\lambda_H}{2}\|H\|_F^2
\label{eq:ufm_centered}
\end{equation}
We denote the empirical covariance by $\Sigma = \frac{1}{M}\tilde{Y}\tilde{Y}^\top \in \mathbb{R}^{n \times n}$, and assume $\Sigma$ is full rank with eigenvalues $\lambda_1 \geq \lambda_2 \geq \cdots \geq \lambda_n > 0$. A key quantity is $c = \lambda_W \lambda_H$.

We first restate the characterization of global minimizers when regularization is present, adapted from Theorem 4.1 of \citet{andriopoulos2024prevalence}. We focus on the regime where $c < \lambda_n$ since we are interested in the limiting behavior when $\lambda_H \to 0$ and $\lambda_W \to 0$.

\begin{theorem}[Regularized UFM Solution, adapted from \citet{andriopoulos2024prevalence}]
\label{thm:ufm_regularized}
Suppose $0 < c < \lambda_n$. Define $A = \Sigma^{1/2} - \sqrt{c} I_n$. Then any global minimizer $(W_\lambda, H_\lambda)$ of \eqref{eq:ufm_centered} can be expressed as:
\begin{align}
W_\lambda &= \left(\frac{\lambda_H}{\lambda_W}\right)^{1/4} A^{1/2} R \label{eq:w_lambda}\\
H_\lambda &= \left(\frac{\lambda_W}{\lambda_H}\right)^{1/4} R^\top A^{1/2} (\Sigma^{1/2})^{-1} \tilde{Y} \label{eq:h_lambda}
\end{align}
where $R \in \mathbb{R}^{n \times d}$ is any matrix satisfying $RR^\top = I_n$.
\end{theorem}

\begin{proof}
This follows from Theorem 4.1 of \citet{andriopoulos2024prevalence} with $c < \lambda_n$ implying $j^* = n$.
\end{proof}

The regularized solution has a specific dimensional structure. Since $A^{1/2}(\Sigma^{1/2})^{-1}\tilde{Y} \in \mathbb{R}^{n \times M}$ and $R^\top \in \mathbb{R}^{d \times n}$, the matrix $H_\lambda \in \mathbb{R}^{d \times M}$ has columns lying in the column space of $R^\top$, which is at most $n$-dimensional. Even though $H_\lambda$ lives in a $d$-dimensional ambient space with $d \gg n$, its columns are confined to an $n$-dimensional subspace.

When regularization is absent, the problem has infinitely many global minimizers, characterized by the following theorem from \citet{andriopoulos2024prevalence}.

\begin{theorem}[Unregularized UFM Solutions, from \citet{andriopoulos2024prevalence}]
\label{thm:ufm_unregularized}
When $\lambda_W = \lambda_H = 0$, a pair $(W, H)$ is a global minimizer of \eqref{eq:ufm_centered} if and only if $WH = \tilde{Y}$ and $W$ has full row rank. For any such $W$, the corresponding global minimizers in $H$ are:
\begin{equation}
H_{\text{unreg}} = W^\dagger \tilde{Y} + (I_d - W^\dagger W)Z
\label{eq:h_unreg}
\end{equation}
where $W^\dagger$ is the Moore--Penrose pseudoinverse and $Z \in \mathbb{R}^{d \times M}$ is arbitrary.
\end{theorem}

\begin{proof}
See Theorem 4.3 of \citet{andriopoulos2024prevalence}.
\end{proof}

The structure in \eqref{eq:h_unreg} decomposes solutions into two orthogonal components: $W^\dagger \tilde{Y}$ lies in the row space of $W$ (dimension at most $n$), while $(I_d - W^\dagger W)Z$ lies in the null space of $W$ (dimension exactly $d - n$). The Frobenius norm satisfies $\|H_{\text{unreg}}\|_F^2 = \|W^\dagger \tilde{Y}\|_F^2 + \|(I_d - W^\dagger W)Z\|_F^2$ by orthogonality. The minimum-norm solution is achieved when $Z = 0$, eliminating the $(d-n)$-dimensional null space component. When $Z \neq 0$, the null space allows $H$ to span up to $d$ dimensions, whereas $Z = 0$ confines $H$ to at most $n$ dimensions.

We now investigate what happens to the regularized solution as weight decay vanishes, examining the limit $\lambda_W, \lambda_H \to 0$ while maintaining $\lambda_H / \lambda_W = k > 0$.

\begin{lemma}[Limiting Reconstruction]
\label{lem:reconstruction_limit}
Suppose $\lim_{\lambda_H \to 0, \lambda_W \to 0}(\lambda_H/\lambda_W) = k > 0$ and let $(W_\lambda, H_\lambda)$ be as in Theorem~\ref{thm:ufm_regularized}. Then:
\begin{equation}
\lim_{\lambda_W, \lambda_H \to 0} W_\lambda H_\lambda = \tilde{Y}
\end{equation}
\end{lemma}

\begin{proof}
Since $c = \lambda_W \lambda_H \to 0$, Theorem~\ref{thm:ufm_regularized} applies for sufficiently small $\lambda_W, \lambda_H$. Multiplying:
\begin{align}
W_\lambda H_\lambda &= \left(\frac{\lambda_H}{\lambda_W}\right)^{1/4} A^{1/2} R \cdot \left(\frac{\lambda_W}{\lambda_H}\right)^{1/4} R^\top A^{1/2} (\Sigma^{1/2})^{-1} \tilde{Y}\\
&= A^{1/2} (RR^\top) A^{1/2} (\Sigma^{1/2})^{-1} \tilde{Y} = A (\Sigma^{1/2})^{-1} \tilde{Y}
\end{align}
Substituting $A = \Sigma^{1/2} - \sqrt{c}I_n$:
\begin{equation}
W_\lambda H_\lambda = (\Sigma^{1/2} - \sqrt{c}I_n)(\Sigma^{1/2})^{-1}\tilde{Y} = (I_n - \sqrt{c}\Sigma^{-1/2})\tilde{Y}
\end{equation}
As $c \to 0$, this yields $\tilde{Y}$.
\end{proof}

\begin{theorem}[Limiting Solution Structure]
\label{thm:limiting_structure}
Under the assumptions of Lemma~\ref{lem:reconstruction_limit}, define:
\begin{align}
W_0 &= \lim_{\lambda_W, \lambda_H \to 0} W_\lambda = k^{1/4} \Sigma^{1/4} R\\
H_0 &= \lim_{\lambda_W, \lambda_H \to 0} H_\lambda = k^{-1/4} R^\top \Sigma^{-1/4} \tilde{Y}
\end{align}
Then $(W_0, H_0)$ is a global minimizer of the unregularized problem with:
\begin{equation}
H_0 = W_0^\dagger \tilde{Y}
\end{equation}
In particular, $H_0$ has no null space component (i.e., $Z = 0$ in Theorem~\ref{thm:ufm_unregularized}).
\end{theorem}

\begin{proof}
From Lemma~\ref{lem:reconstruction_limit}, $W_0 H_0 = \tilde{Y}$, so $(W_0, H_0)$ is a global minimizer. Since $W_0 = k^{1/4}\Sigma^{1/4}R$ has full row rank, its pseudoinverse is:
\begin{equation}
W_0^\dagger = W_0^\top (W_0 W_0^\top)^{-1} = (k^{1/4}\Sigma^{1/4}R)^\top (k^{1/2}\Sigma^{1/2})^{-1} = k^{-1/4}R^\top \Sigma^{-1/4}
\end{equation}
Thus $W_0^\dagger \tilde{Y} = k^{-1/4} R^\top \Sigma^{-1/4} \tilde{Y} = H_0$, confirming $Z = 0$. For any $H$ satisfying $W_0 H = \tilde{Y}$, we have $H = W_0^\dagger \tilde{Y} + (I_d - W_0^\dagger W_0)Z$ with:
\begin{equation}
\|H\|_F^2 = \|W_0^\dagger \tilde{Y}\|_F^2 + \|(I_d - W_0^\dagger W_0)Z\|_F^2 \geq \|H_0\|_F^2
\end{equation}
by orthogonality, with equality if and only if $Z = 0$. Thus $H_0$ is the minimum-norm solution.
\end{proof}

These results explain why weight decay causes dimensional collapse. Without regularization, $H$ can utilize the full $d$-dimensional space through the arbitrary null space component $(I_d - W^\dagger W)Z$, where the null space has dimension $d - n$. With any positive regularization, Theorem~\ref{thm:ufm_regularized} shows $H_\lambda$ is confined to an at most $n$-dimensional subspace. Theorem~\ref{thm:limiting_structure} proves that even as $\lambda_W, \lambda_H \to 0^+$, the limiting solution has $Z = 0$, eliminating the $(d-n)$-dimensional null space component. This demonstrates that even infinitesimally small weight decay induces dimensional collapse by biasing toward the minimum-norm solution of the unregularized problem, which lies entirely in the $n$-dimensional row space of $W$. Since typically $d \gg n$, this represents a massive dimensional reduction from the ambient feature space to a low-dimensional subspace determined by the target structure.

\begin{remark}[Breakdown of Affine Congruence in the Over-Compressed Regime.]
\label{rem:ufm_affine_congruency}

Theorem~\ref{thm:ufm_regularized}, adapted from Theorem~4.1 of \citet{andriopoulos2024prevalence}, characterizes global minimizers of the regularized UFM objective in terms of the parameter
\[
c := \lambda_W \lambda_H
\]
and the eigenvalues $\lambda_1 \ge \cdots \ge \lambda_n > 0$ of the target covariance $\Sigma$.
When $0 < c < \lambda_n$, the theorem yields $j^{\ast} = n$, and the solution satisfies
\[
H_\lambda
= \left(\frac{\lambda_W}{\lambda_H}\right)^{1/4}
R^\top A^{1/2} (\Sigma^{1/2})^{-1} \tilde{Y},
\quad
A = \Sigma^{1/2} - \sqrt{c}\, I_n,
\]
with $A \succ 0$. In this regime, $A^{1/2}$ is full rank and the map relating $H_\lambda$ and $\tilde{Y}$ is an invertible affine transformation on the support of $\tilde{Y}$. Consequently, the learned features and targets are affine congruent, implying preservation of intrinsic dimension:
\[
\idh = \idy.
\]

In contrast, when $c \ge \lambda_n$, Theorem~4.1 of \citet{andriopoulos2024prevalence} yields $j^{\ast} < n$, and the solution involves truncated matrices $[A^{1/2}]_{j^{\ast}}$ and $[\Sigma^{1/2}]_{j^{\ast}}$ obtained by retaining only the eigenspaces corresponding to eigenvalues $\lambda_i > c$. In this case, the resulting feature matrix satisfies
\[
\text{rank}(H_\lambda) \le j^{\ast} < n,
\]
and the affine map relating $H_\lambda$ and $\tilde{Y}$ is no longer invertible on the support of $\tilde{Y}$. As a consequence, affine congruency is not guaranteed to hold. This regime allows for a reduction in intrinsic dimension,
\[
\idh < \idy,
\]
which is precisely what we observe empirically in the over-compressed region. This observation highlights the diagnostic value of intrinsic dimension: deviations between $\idh$ and $\idy$ signal departure from the regime in which the assumptions underlying affine congruency hold. Although the parameter $c$ is not directly observable in practice, as it arises from the UFM abstraction rather than explicit training dynamics, intrinsic dimension provides an empirical proxy for this regime. In particular, the transition from $\idh \approx \idy$ to $\idh < \idy$ indicates that the effective regularization has crossed the threshold $c = \lambda_n$, placing the model in the over-compressed regime and violating the conditions under which affine congruency is expected.

\end{remark}

\subsection{Why Collapsed Models Fail to Generalize?}
\label{appx:non_surjectivity}
The proof of Theorem~\ref{thm:non_surjectivity} follows directly from Sard's theorem. 

\begin{theorem}
Let $\mathcal{M}$ be a smooth $m$-dimensional manifold and $\mathcal{N}$ be a smooth $n$-dimensional manifold, with $m < n$. A smooth map $g: \mathcal{M} \to \mathcal{N}$ cannot be surjective, i.e., $g(\mathcal{M}) \neq \mathcal{N}$.
\end{theorem}

\begin{proof}
Let $g: \mathcal{M} \to \mathcal{N}$ be a smooth map where $\dim(\mathcal{M}) = m$ and $\dim(\mathcal{N}) = n$, under the condition $m < n$. Consider an arbitrary point $p \in \mathcal{M}$. The differential of the map at this point, $dg_p: T_p\mathcal{M} \to T_{g(p)}\mathcal{N}$, is a linear transformation from the $m$-dimensional tangent space of $\mathcal{M}$ at $p$ to the $n$-dimensional tangent space of $\mathcal{N}$ at $g(p)$.

By the rank-nullity theorem, the rank of $dg_p$ is bounded by the dimension of its domain, so it holds that $\text{rank}(dg_p) \le m$. Given that $m < n$, it follows that $\text{rank}(dg_p) < n$. A linear map is surjective if and only if its rank equals the dimension of its codomain; thus, $dg_p$ is not surjective.

As the choice of $p$ was arbitrary, this holds for all $p \in \mathcal{M}$. By definition, a point is critical if its differential is not surjective. Therefore, every point in the domain $\mathcal{M}$ is a critical point of $g$. The image of the set of critical points is the set of {critical values}. In this case, the set of critical values is the entire image of the map, $g(\mathcal{M})$.

By Sard's Theorem, the set of critical values of a smooth map has Lebesgue measure zero in the codomain. It follows that the image $g(\mathcal{M})$ has measure zero in $\mathcal{N}$. However, a smooth $n$-dimensional manifold (for $n \ge 1$) has positive Lebesgue measure. Since a set of measure zero cannot be equal to a set of positive measure, it must be that $g(\mathcal{M}) \neq \mathcal{N}$.

Therefore, the map $g$ is not surjective.
\end{proof}

This theorem provides the geometric foundation for understanding the failure of collapsed models. In our regression context, the learned features $\{\bh_\theta(\bx)\}$ form a feature manifold $\mathcal{M}_H$ of dimension $m = \idh$, while the targets $\{\by\}$ lie on a target manifold $\mathcal{N}_Y$ of dimension $n = \idy$. The final layer of the network constitutes a smooth map from the feature manifold to the target space.

When a model is in the over-compressed regime ($\idh < \idy$), the theorem's condition ($m < n$) is met. The direct consequence is that this smooth map cannot be surjective. This means the image of the feature manifold---the set of all possible predictions the model can generate---is a proper subset of the target manifold. Geometrically, there will always be points on the target manifold that lie outside the model's predictive reach. A perfect reconstruction is therefore impossible, as the model is fundamentally incapable of generating the full range of target data, leading to an unavoidable error.

\newpage
\section*{NeurIPS Paper Checklist}

\begin{enumerate}

\item {\bf Claims}
    \item[] Question: Do the main claims made in the abstract and introduction accurately reflect the paper's contributions and scope?
    \item[] Answer: \answerYes{} 
    \item[] Justification: The three contributions listed in the introduction are directly supported by Section~\ref{sec:nrc_geometry}, Sections~\ref{sec:generalization}--\ref{sec:mathematical_analysis}, and Section~\ref{sec:practical}, respectively.
    \item[] Guidelines:
    \begin{itemize}
        \item The answer \answerNA{} means that the abstract and introduction do not include the claims made in the paper.
        \item The abstract and/or introduction should clearly state the claims made, including the contributions made in the paper and important assumptions and limitations. A \answerNo{} or \answerNA{} answer to this question will not be perceived well by the reviewers. 
        \item The claims made should match theoretical and experimental results, and reflect how much the results can be expected to generalize to other settings. 
        \item It is fine to include aspirational goals as motivation as long as it is clear that these goals are not attained by the paper. 
    \end{itemize}

\item {\bf Limitations}
    \item[] Question: Does the paper discuss the limitations of the work performed by the authors?
    \item[] Answer: \answerYes{} 
    \item[] Justification: Section~\ref{sec:limitation} provides a detailed discussion of limitations and outlines directions for future work.
    \item[] Guidelines:
    \begin{itemize}
        \item The answer \answerNA{} means that the paper has no limitation while the answer \answerNo{} means that the paper has limitations, but those are not discussed in the paper. 
        \item The authors are encouraged to create a separate ``Limitations'' section in their paper.
        \item The paper should point out any strong assumptions and how robust the results are to violations of these assumptions (e.g., independence assumptions, noiseless settings, model well-specification, asymptotic approximations only holding locally). The authors should reflect on how these assumptions might be violated in practice and what the implications would be.
        \item The authors should reflect on the scope of the claims made, e.g., if the approach was only tested on a few datasets or with a few runs. In general, empirical results often depend on implicit assumptions, which should be articulated.
        \item The authors should reflect on the factors that influence the performance of the approach. For example, a facial recognition algorithm may perform poorly when image resolution is low or images are taken in low lighting. Or a speech-to-text system might not be used reliably to provide closed captions for online lectures because it fails to handle technical jargon.
        \item The authors should discuss the computational efficiency of the proposed algorithms and how they scale with dataset size.
        \item If applicable, the authors should discuss possible limitations of their approach to address problems of privacy and fairness.
        \item While the authors might fear that complete honesty about limitations might be used by reviewers as grounds for rejection, a worse outcome might be that reviewers discover limitations that aren't acknowledged in the paper. The authors should use their best judgment and recognize that individual actions in favor of transparency play an important role in developing norms that preserve the integrity of the community. Reviewers will be specifically instructed to not penalize honesty concerning limitations.
    \end{itemize}

\item {\bf Theory assumptions and proofs}
    \item[] Question: For each theoretical result, does the paper provide the full set of assumptions and a complete (and correct) proof?
    \item[] Answer: \answerYes{} 
    \item[] Justification: Appendix~\ref{appx:mathematical_analysis} includes full details about our theoretical results.
    \item[] Guidelines:
    \begin{itemize}
        \item The answer \answerNA{} means that the paper does not include theoretical results. 
        \item All the theorems, formulas, and proofs in the paper should be numbered and cross-referenced.
        \item All assumptions should be clearly stated or referenced in the statement of any theorems.
        \item The proofs can either appear in the main paper or the supplemental material, but if they appear in the supplemental material, the authors are encouraged to provide a short proof sketch to provide intuition. 
        \item Inversely, any informal proof provided in the core of the paper should be complemented by formal proofs provided in appendix or supplemental material.
        \item Theorems and Lemmas that the proof relies upon should be properly referenced. 
    \end{itemize}

    \item {\bf Experimental result reproducibility}
    \item[] Question: Does the paper fully disclose all the information needed to reproduce the main experimental results of the paper to the extent that it affects the main claims and/or conclusions of the paper (regardless of whether the code and data are provided or not)?
    \item[] Answer: \answerYes{} 
    \item[] Justification: Appendix \ref{appx:exp_details} lists all experimental details for reproduction.
    \item[] Guidelines:
    \begin{itemize}
        \item The answer \answerNA{} means that the paper does not include experiments.
        \item If the paper includes experiments, a \answerNo{} answer to this question will not be perceived well by the reviewers: Making the paper reproducible is important, regardless of whether the code and data are provided or not.
        \item If the contribution is a dataset and\slash or model, the authors should describe the steps taken to make their results reproducible or verifiable. 
        \item Depending on the contribution, reproducibility can be accomplished in various ways. For example, if the contribution is a novel architecture, describing the architecture fully might suffice, or if the contribution is a specific model and empirical evaluation, it may be necessary to either make it possible for others to replicate the model with the same dataset, or provide access to the model. In general. releasing code and data is often one good way to accomplish this, but reproducibility can also be provided via detailed instructions for how to replicate the results, access to a hosted model (e.g., in the case of a large language model), releasing of a model checkpoint, or other means that are appropriate to the research performed.
        \item While NeurIPS does not require releasing code, the conference does require all submissions to provide some reasonable avenue for reproducibility, which may depend on the nature of the contribution. For example
        \begin{enumerate}
            \item If the contribution is primarily a new algorithm, the paper should make it clear how to reproduce that algorithm.
            \item If the contribution is primarily a new model architecture, the paper should describe the architecture clearly and fully.
            \item If the contribution is a new model (e.g., a large language model), then there should either be a way to access this model for reproducing the results or a way to reproduce the model (e.g., with an open-source dataset or instructions for how to construct the dataset).
            \item We recognize that reproducibility may be tricky in some cases, in which case authors are welcome to describe the particular way they provide for reproducibility. In the case of closed-source models, it may be that access to the model is limited in some way (e.g., to registered users), but it should be possible for other researchers to have some path to reproducing or verifying the results.
        \end{enumerate}
    \end{itemize}

\item {\bf Open access to data and code}
    \item[] Question: Does the paper provide open access to the data and code, with sufficient instructions to faithfully reproduce the main experimental results, as described in supplemental material?
    \item[] Answer: \answerYes{} 
    \item[] Justification: All datasets are publicly available, and our code is released at \url{https://github.com/Konokiii/neural_regression_collapse}.

    \item[] Guidelines:
    \begin{itemize}
        \item The answer \answerNA{} means that paper does not include experiments requiring code.
        \item Please see the NeurIPS code and data submission guidelines (\url{https://neurips.cc/public/guides/CodeSubmissionPolicy}) for more details.
        \item While we encourage the release of code and data, we understand that this might not be possible, so \answerNo{} is an acceptable answer. Papers cannot be rejected simply for not including code, unless this is central to the contribution (e.g., for a new open-source benchmark).
        \item The instructions should contain the exact command and environment needed to run to reproduce the results. See the NeurIPS code and data submission guidelines (\url{https://neurips.cc/public/guides/CodeSubmissionPolicy}) for more details.
        \item The authors should provide instructions on data access and preparation, including how to access the raw data, preprocessed data, intermediate data, and generated data, etc.
        \item The authors should provide scripts to reproduce all experimental results for the new proposed method and baselines. If only a subset of experiments are reproducible, they should state which ones are omitted from the script and why.
        \item At submission time, to preserve anonymity, the authors should release anonymized versions (if applicable).
        \item Providing as much information as possible in supplemental material (appended to the paper) is recommended, but including URLs to data and code is permitted.
    \end{itemize}

\item {\bf Experimental setting/details}
    \item[] Question: Does the paper specify all the training and test details (e.g., data splits, hyperparameters, how they were chosen, type of optimizer) necessary to understand the results?
    \item[] Answer: \answerYes{} 
    \item[] Justification: Full training and evaluation details, including all hyperparameter grids, are provided in Appendix~\ref{appx:exp_details} and summarized in Tables~\ref{table:mujoco_params} and~\ref{table:mnist_cifar_params}.
    \item[] Guidelines:
    \begin{itemize}
        \item The answer \answerNA{} means that the paper does not include experiments.
        \item The experimental setting should be presented in the core of the paper to a level of detail that is necessary to appreciate the results and make sense of them.
        \item The full details can be provided either with the code, in appendix, or as supplemental material.
    \end{itemize}

\item {\bf Experiment statistical significance}
    \item[] Question: Does the paper report error bars suitably and correctly defined or other appropriate information about the statistical significance of the experiments?
    \item[] Answer: \answerYes{} 
    \item[] Justification: We sweep over model widths and weight decay values to demonstrate robustness of trends, and RL policy evaluations are averaged over 100 episodes.
    \item[] Guidelines:
    \begin{itemize}
        \item The answer \answerNA{} means that the paper does not include experiments.
        \item The authors should answer \answerYes{} if the results are accompanied by error bars, confidence intervals, or statistical significance tests, at least for the experiments that support the main claims of the paper.
        \item The factors of variability that the error bars are capturing should be clearly stated (for example, train/test split, initialization, random drawing of some parameter, or overall run with given experimental conditions).
        \item The method for calculating the error bars should be explained (closed form formula, call to a library function, bootstrap, etc.)
        \item The assumptions made should be given (e.g., Normally distributed errors).
        \item It should be clear whether the error bar is the standard deviation or the standard error of the mean.
        \item It is OK to report 1-sigma error bars, but one should state it. The authors should preferably report a 2-sigma error bar than state that they have a 96\% CI, if the hypothesis of Normality of errors is not verified.
        \item For asymmetric distributions, the authors should be careful not to show in tables or figures symmetric error bars that would yield results that are out of range (e.g., negative error rates).
        \item If error bars are reported in tables or plots, the authors should explain in the text how they were calculated and reference the corresponding figures or tables in the text.
    \end{itemize}

\item {\bf Experiments compute resources}
    \item[] Question: For each experiment, does the paper provide sufficient information on the computer resources (type of compute workers, memory, time of execution) needed to reproduce the experiments?
    \item[] Answer: \answerYes{} 
    \item[] Justification: Appendix~\ref{appx:exp_details} reports the GPU type, memory, number of CPU workers, and average training time per model.
    \item[] Guidelines:
    \begin{itemize}
        \item The answer \answerNA{} means that the paper does not include experiments.
        \item The paper should indicate the type of compute workers CPU or GPU, internal cluster, or cloud provider, including relevant memory and storage.
        \item The paper should provide the amount of compute required for each of the individual experimental runs as well as estimate the total compute. 
        \item The paper should disclose whether the full research project required more compute than the experiments reported in the paper (e.g., preliminary or failed experiments that didn't make it into the paper). 
    \end{itemize}
    
\item {\bf Code of ethics}
    \item[] Question: Does the research conducted in the paper conform, in every respect, with the NeurIPS Code of Ethics \url{https://neurips.cc/public/EthicsGuidelines}?
    \item[] Answer: \answerYes{} 
    \item[] Justification: We have reviewed and adhered to the NeurIPS Code of Ethics throughout this work.
    \item[] Guidelines:
    \begin{itemize}
        \item The answer \answerNA{} means that the authors have not reviewed the NeurIPS Code of Ethics.
        \item If the authors answer \answerNo, they should explain the special circumstances that require a deviation from the Code of Ethics.
        \item The authors should make sure to preserve anonymity (e.g., if there is a special consideration due to laws or regulations in their jurisdiction).
    \end{itemize}

\item {\bf Broader impacts}
    \item[] Question: Does the paper discuss both potential positive societal impacts and negative societal impacts of the work performed?
    \item[] Answer: \answerNA{} 
    \item[] Justification: This work is foundational research on the geometry of learned representations, with no direct pathway to negative societal applications.
    \item[] Guidelines:
    \begin{itemize}
        \item The answer \answerNA{} means that there is no societal impact of the work performed.
        \item If the authors answer \answerNA{} or \answerNo, they should explain why their work has no societal impact or why the paper does not address societal impact.
        \item Examples of negative societal impacts include potential malicious or unintended uses (e.g., disinformation, generating fake profiles, surveillance), fairness considerations (e.g., deployment of technologies that could make decisions that unfairly impact specific groups), privacy considerations, and security considerations.
        \item The conference expects that many papers will be foundational research and not tied to particular applications, let alone deployments. However, if there is a direct path to any negative applications, the authors should point it out. For example, it is legitimate to point out that an improvement in the quality of generative models could be used to generate Deepfakes for disinformation. On the other hand, it is not needed to point out that a generic algorithm for optimizing neural networks could enable people to train models that generate Deepfakes faster.
        \item The authors should consider possible harms that could arise when the technology is being used as intended and functioning correctly, harms that could arise when the technology is being used as intended but gives incorrect results, and harms following from (intentional or unintentional) misuse of the technology.
        \item If there are negative societal impacts, the authors could also discuss possible mitigation strategies (e.g., gated release of models, providing defenses in addition to attacks, mechanisms for monitoring misuse, mechanisms to monitor how a system learns from feedback over time, improving the efficiency and accessibility of ML).
    \end{itemize}
    
\item {\bf Safeguards}
    \item[] Question: Does the paper describe safeguards that have been put in place for responsible release of data or models that have a high risk for misuse (e.g., pre-trained language models, image generators, or scraped datasets)?
    \item[] Answer: \answerNA{} 
    \item[] Justification: The paper does not release pretrained models or new datasets that pose misuse risks.
    \item[] Guidelines:
    \begin{itemize}
        \item The answer \answerNA{} means that the paper poses no such risks.
        \item Released models that have a high risk for misuse or dual-use should be released with necessary safeguards to allow for controlled use of the model, for example by requiring that users adhere to usage guidelines or restrictions to access the model or implementing safety filters. 
        \item Datasets that have been scraped from the Internet could pose safety risks. The authors should describe how they avoided releasing unsafe images.
        \item We recognize that providing effective safeguards is challenging, and many papers do not require this, but we encourage authors to take this into account and make a best faith effort.
    \end{itemize}

\item {\bf Licenses for existing assets}
    \item[] Question: Are the creators or original owners of assets (e.g., code, data, models), used in the paper, properly credited and are the license and terms of use explicitly mentioned and properly respected?
    \item[] Answer: \answerYes{} 
    \item[] Justification: All datasets (MuJoCo/Gymnasium, D4RL, Visual D4RL) and pretrained models are properly cited, and each is released under an open-source license.
    \item[] Guidelines:
    \begin{itemize}
        \item The answer \answerNA{} means that the paper does not use existing assets.
        \item The authors should cite the original paper that produced the code package or dataset.
        \item The authors should state which version of the asset is used and, if possible, include a URL.
        \item The name of the license (e.g., CC-BY 4.0) should be included for each asset.
        \item For scraped data from a particular source (e.g., website), the copyright and terms of service of that source should be provided.
        \item If assets are released, the license, copyright information, and terms of use in the package should be provided. For popular datasets, \url{paperswithcode.com/datasets} has curated licenses for some datasets. Their licensing guide can help determine the license of a dataset.
        \item For existing datasets that are re-packaged, both the original license and the license of the derived asset (if it has changed) should be provided.
        \item If this information is not available online, the authors are encouraged to reach out to the asset's creators.
    \end{itemize}

\item {\bf New assets}
    \item[] Question: Are new assets introduced in the paper well documented and is the documentation provided alongside the assets?
    \item[] Answer: \answerNA{} 
    \item[] Justification: This paper does not introduce new datasets, models, or benchmarks.

    \item[] Guidelines:
    \begin{itemize}
        \item The answer \answerNA{} means that the paper does not release new assets.
        \item Researchers should communicate the details of the dataset\slash code\slash model as part of their submissions via structured templates. This includes details about training, license, limitations, etc. 
        \item The paper should discuss whether and how consent was obtained from people whose asset is used.
        \item At submission time, remember to anonymize your assets (if applicable). You can either create an anonymized URL or include an anonymized zip file.
    \end{itemize}

\item {\bf Crowdsourcing and research with human subjects}
    \item[] Question: For crowdsourcing experiments and research with human subjects, does the paper include the full text of instructions given to participants and screenshots, if applicable, as well as details about compensation (if any)? 
    \item[] Answer: \answerNA{} 
    \item[] Justification: This work does not involve crowdsourcing or human subjects.
    \item[] Guidelines:
    \begin{itemize}
        \item The answer \answerNA{} means that the paper does not involve crowdsourcing nor research with human subjects.
        \item Including this information in the supplemental material is fine, but if the main contribution of the paper involves human subjects, then as much detail as possible should be included in the main paper. 
        \item According to the NeurIPS Code of Ethics, workers involved in data collection, curation, or other labor should be paid at least the minimum wage in the country of the data collector. 
    \end{itemize}

\item {\bf Institutional review board (IRB) approvals or equivalent for research with human subjects}
    \item[] Question: Does the paper describe potential risks incurred by study participants, whether such risks were disclosed to the subjects, and whether Institutional Review Board (IRB) approvals (or an equivalent approval/review based on the requirements of your country or institution) were obtained?
    \item[] Answer: \answerNA{} 
    \item[] Justification: This work does not involve human subjects research.
    \item[] Guidelines:
    \begin{itemize}
        \item The answer \answerNA{} means that the paper does not involve crowdsourcing nor research with human subjects.
        \item Depending on the country in which research is conducted, IRB approval (or equivalent) may be required for any human subjects research. If you obtained IRB approval, you should clearly state this in the paper. 
        \item We recognize that the procedures for this may vary significantly between institutions and locations, and we expect authors to adhere to the NeurIPS Code of Ethics and the guidelines for their institution. 
        \item For initial submissions, do not include any information that would break anonymity (if applicable), such as the institution conducting the review.
    \end{itemize}

\item {\bf Declaration of LLM usage}
    \item[] Question: Does the paper describe the usage of LLMs if it is an important, original, or non-standard component of the core methods in this research? Note that if the LLM is used only for writing, editing, or formatting purposes and does \emph{not} impact the core methodology, scientific rigor, or originality of the research, declaration is not required.
    \item[] Answer: \answerNA{} 
    \item[] Justification: LLMs were used solely for writing assistance and editing; they played no role in the core methodology or experiments.
    \item[] Guidelines:
    \begin{itemize}
        \item The answer \answerNA{} means that the core method development in this research does not involve LLMs as any important, original, or non-standard components.
        \item Please refer to our LLM policy in the NeurIPS handbook for what should or should not be described.
    \end{itemize}

\end{enumerate}

\end{document}